\documentclass[letterpaper]{article}
\usepackage[preprint]{aaai2027}

\usepackage[hyphens]{url}
\usepackage{graphicx}
\def\UrlFont{\rm}
\usepackage{natbib}
\usepackage{caption}
\usepackage{booktabs}
\usepackage{multirow}
\usepackage[table]{xcolor}
\usepackage{adjustbox}
\usepackage{amsmath,amssymb,mathtools}
\usepackage{algorithm}
\usepackage[noend]{algpseudocode}
\usepackage{dblfloatfix}

\usepackage{etoolbox}
\makeatletter
\newcommand{\aaaiteaser}{%
  \centering
  \vspace{-18mm}
  \includegraphics[width=\textwidth]{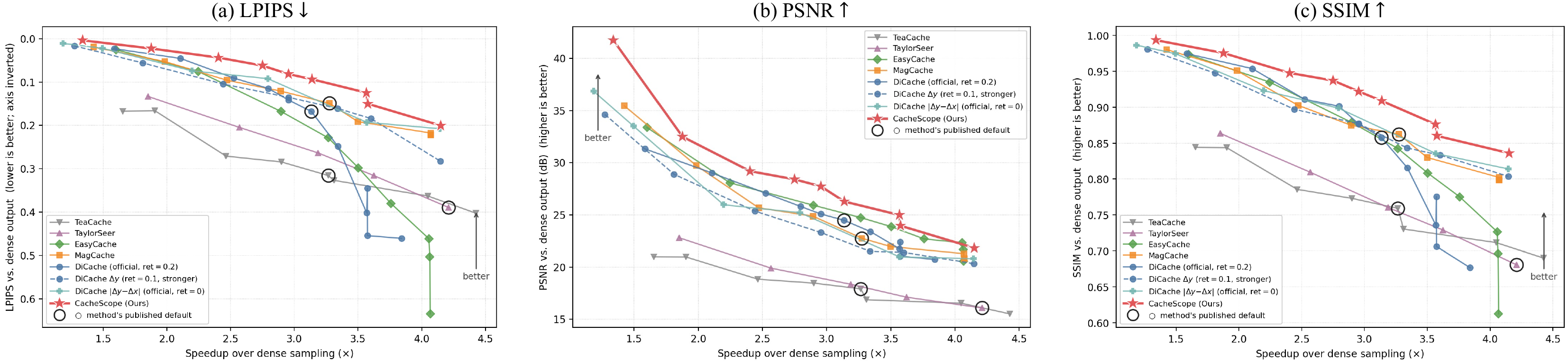}%
  \vspace{-2mm}
  \captionof{figure}
  {\textbf{Quality--speedup trade-off on FLUX.1-dev.}
  Median (a) LPIPS, (b) PSNR, and (c) SSIM against the dense
  (full-computation) output as a function of measured end-to-end
  speedup, for CacheScope and training-free caching baselines;
  circled markers denote each method's published default
  configuration. CacheScope (red) traces the best frontier on all
  three metrics across the entire acceleration range.\label{fig:teaser}}%
  \vspace{2pt}%
}
\AtBeginDocument{%
  \patchcmd{\maketitle}{\twocolumn[\@maketitle]}%
    {\twocolumn[{\@maketitle\aaaiteaser}]}{}{%
    \PackageWarning{AAmain}{teaser patch failed}}%
}
\makeatother
\usepackage{needspace}
\usepackage{tikz}
\usetikzlibrary{arrows.meta,positioning}

\usepackage{booktabs}
\usepackage{multirow}
\usepackage{tabularx}
\usepackage{array}
\usepackage[table]{xcolor}
\definecolor{lightblue}{RGB}{198,226,255}
\usepackage{tcolorbox}
\title{CacheScope: Exact-Compute Allocation for Training-Free Diffusion Acceleration}

\author{
    Qinhan Lyu, Jusheng Zhang, Zhixing Zhang, Hao Liu, Zijie Liang, Jin Jiang, Keze Wang\corresponding
}

\affiliations{
    Sun Yat-sen University
}

\begin{document}
\maketitle

\begin{abstract}
Training-free feature reuse accelerates diffusion transformer inference by replacing full denoising computation at selected steps with cached approximations. Existing methods mainly improve the reconstruction of skipped features, but we find that once a stable first-order reconstructor is used, increasing reconstruction complexity brings limited gains, whereas \textbf{the placement of exact-computation steps has a much larger effect on fidelity}. More importantly, \textbf{absolute shallow-probe variation is jointly affected by input-trajectory displacement and network response strength, and therefore does not directly reflect cache risk}. We address this confounding by normalizing probe-output variation with the corresponding input displacement, yielding a directional gain per unit trajectory movement. Based on this signal, we propose \textbf{CacheScope}, which identifies high-risk reuse steps online and selectively falls back to full denoising computation while leaving the underlying cache reconstructor unchanged. CacheScope requires no additional training, gradients, auxiliary networks, or prompt-specific offline schedule optimization. Across FLUX.1-dev, Wan2.1-T2V-1.3B, and Open-Sora 1.2, CacheScope achieves up to $3.57\times$ acceleration and reduces median LPIPS by up to $70.1\%$ over the strongest probe-based baseline.
\end{abstract}

\section{Introduction}

Diffusion and flow-based generative models~\cite{ddpm,score_sde,ldm,flow_matching,rectified_flow} have become a dominant paradigm for high-fidelity image and video synthesis~\cite{videodm,cogvideox,hunyuanvideo,wan,wang2025scalingdeepfake}. Related efforts also target controllable video generation and embodied action models~\cite{coagent,e0,zhang2026actphyscause,zhang2025osc,zhang2025kabb}. Their generation process, however, requires repeatedly evaluating a large denoising network along a discretized trajectory. For modern diffusion transformers (DiTs)~\cite{dit}, each evaluation performs expensive attention and feed-forward computation over long spatial or spatiotemporal token sequences, making inference latency and computational cost remain substantial. Training-free feature reuse provides a practical acceleration route by replacing selected denoising evaluations with inexpensive approximations constructed from previously cached states. Since it requires neither retraining nor modification of the pretrained model weights, feature reuse can be directly applied to rapidly evolving diffusion backbones.

A training-free reuse method must answer two fundamental questions: \emph{how to reuse} previously computed information, and \emph{when to reuse} it. The former determines how the output of a skipped denoising evaluation is reconstructed, while the latter determines which steps should retain exact computation. Both questions are important. A weak reconstruction rule can immediately introduce visible errors and cause deviations to accumulate along the generation trajectory. Consequently, a large body of recent work has explored increasingly sophisticated reconstruction strategies, ranging from direct feature reuse and scalar correction to trajectory extrapolation and multi-history regression~\cite{deepcache,teacache,magcache,dicache}.

\begin{figure*}[t]
    \centering
    \includegraphics[width=\textwidth]{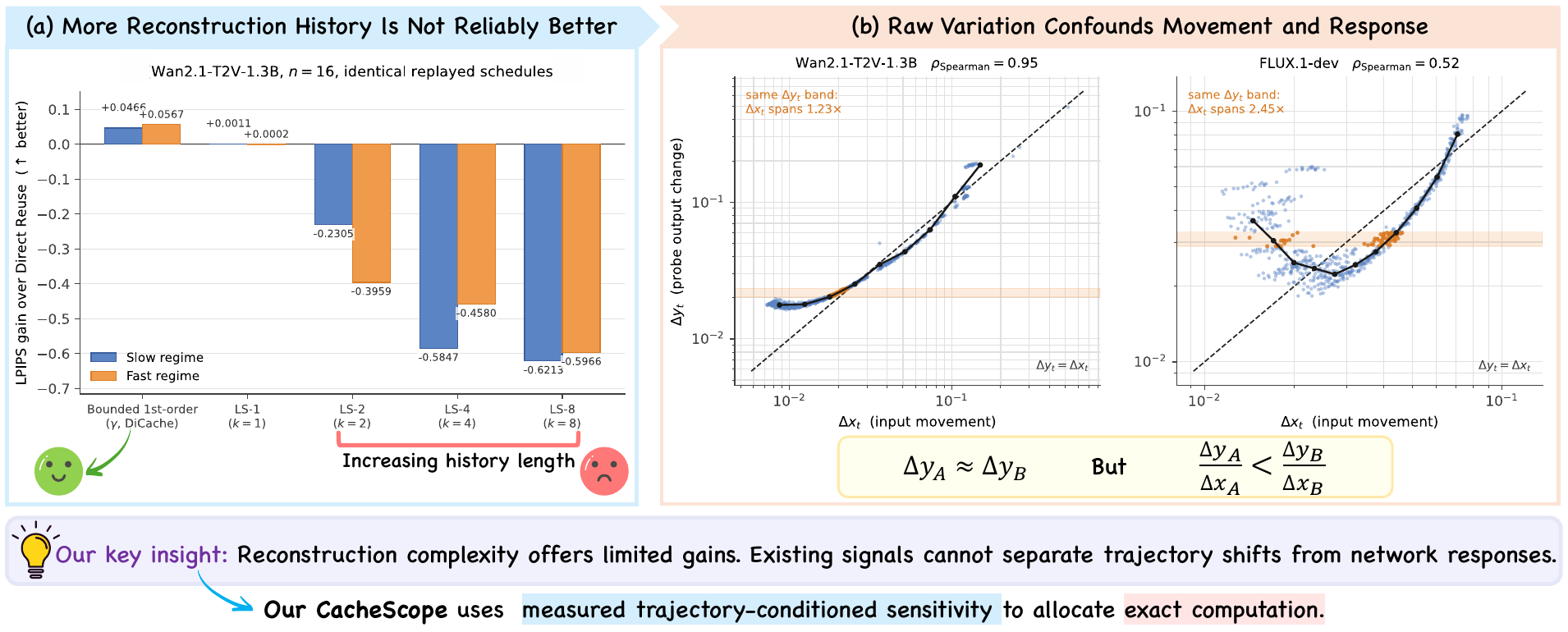}
     \vspace{-5mm}
    \caption{\textbf{Overview of the CacheScope insight.} Under a fixed cache
    reconstructor and a fixed exact-compute budget, more flexible
    reconstruction rules do not consistently outperform a bounded
    first-order reconstructor, while
    the placement of exact evaluations dominates fidelity. Raw probe
    variation $\Delta y_t$ conflates trajectory displacement with network
    response; CacheScope instead schedules exact computation using the
    trajectory-conditioned sensitivity $s_t=\Delta y_t/(\Delta x_t+\epsilon)$.}
    \label{fig:insight}
    \vspace{-4mm}
\end{figure*}

Nevertheless, our controlled experiments reveal a clear asymmetry between these two design dimensions. Once a stable bounded first-order reconstructor is available, the more flexible reconstruction formulations we examine do not consistently outperform it. Extending the residual history, introducing higher-order fitting rules, or using more flexible regression formulations improves fidelity in some settings but degrades it in others. On video diffusion transformers, unconstrained multi-history reconstruction can even become unstable and perform substantially worse than a simple first-order alternative. We therefore do not argue that reconstruction is unimportant, nor that no better reconstructor exists; rather, within the examined design space, we did not find a formulation that consistently outperforms the lightweight first-order rule.

In contrast, we find that \textbf{the placement of exact computation remains highly consequential}: even when the cache reconstructor, probe depth, warm-up policy, and total number of exact denoiser evaluations are all fixed, different reuse schedules can produce markedly different generation quality. Thus, the central bottleneck is not merely whether a sufficiently strong reconstruction rule is used, but whether a limited exact-compute budget is allocated to the right locations along the denoising trajectory. This observation shifts the focus of diffusion caching from designing increasingly elaborate approximators to a more fundamental allocation problem:
\begin{quote}
\emph{Under a fixed exact-computation budget, which denoising steps are most valuable to compute exactly?}
\end{quote}

Existing adaptive schedulers commonly address this problem by measuring the raw variation of a shallow probe output. Intuitively, a large probe change is interpreted as evidence that the current step is difficult to approximate and should therefore receive an exact forward pass. However, this interpretation implicitly assumes that transitions at different denoising steps are directly comparable. We find that this assumption does not hold along diffusion trajectories: \textbf{the observed probe-output variation is jointly determined by two factors---how far the input trajectory moves between adjacent steps, and how strongly the network responds to that movement} (Figure~\ref{fig:insight}). A large output change may therefore arise simply because the latent input has moved a long distance, rather than because the network is locally sensitive. Conversely, a small trajectory movement can trigger a disproportionately strong network response while still producing only a moderate absolute variation.

This distinction can be seen through a local first-order view. Let $x_t$ denote the shallow-probe input, $y_t=g(x_t)$ its output, and $J_t$ the local Jacobian of the probe. Then
\begin{equation}
y_t-y_{t-1}
\approx
J_t(x_t-x_{t-1}).
\end{equation}
The magnitude of the observed output change therefore entangles the trajectory displacement $x_t-x_{t-1}$ with the directional network response induced by that displacement. Raw probe variation alone cannot reliably distinguish a large but proportionate response from a small movement that is strongly amplified by the model. As a result, it is not a directly comparable measure of the value of exact computation across denoising steps---that is, it does not directly reflect the risk of reusing cache at a given step.

Motivated by this insight, we introduce \textbf{CacheScope}, a lightweight drop-in scheduling rule for online exact-compute allocation. Rather than ranking steps by absolute probe variation, CacheScope measures the network response relative to the corresponding local trajectory movement. Specifically, it constructs a trajectory-conditioned sensitivity score from the ratio between relative probe-output variation and relative probe-input movement:
\begin{equation}
s_t
=
\frac{\Delta y_t}{\Delta x_t+\epsilon}.
\end{equation}
This response-per-movement quantity serves as a lightweight proxy for the local directional sensitivity of the denoising network. CacheScope accumulates the sensitivity score online and performs an exact denoiser evaluation when the accumulated utility reaches a predefined threshold; otherwise, the step is approximated using the existing cache reconstructor. In this way, CacheScope converts trajectory-aware sensitivity estimation into a budgeted allocation policy for exact computation.

Importantly, CacheScope changes the semantics of the scheduling decision without increasing the complexity of the reuse mechanism. It requires no training, gradients, auxiliary prediction network, online optimization, or offline schedule search. It leaves the underlying cache reconstructor and cached history unchanged, and introduces only negligible scalar computation beyond an existing shallow probe. CacheScope can therefore be incorporated into existing probe-based caching pipelines without altering their reconstruction components.

We evaluate CacheScope on the image diffusion transformer FLUX.1-dev and the video diffusion transformers Wan2.1-T2V-1.3B and Open-Sora 1.2. Under strictly matched exact-computation budgets, CacheScope consistently improves perceptual fidelity across all backbones and multiple acceleration regimes. It establishes a new end-to-end speed--fidelity frontier among the evaluated training-free feature-reuse methods (Figure~\ref{fig:teaser}), achieving up to $3.57\times$ acceleration and reducing median LPIPS by up to $70.1\%$ over the strongest probe-based baseline. At about $1.40\times$ acceleration on Open-Sora 1.2, CacheScope reduces LPIPS from $0.0566$ to $0.0169$; at about $2.95\times$ on FLUX, from $0.1353$ to $0.0816$; and at about $3.46\times$ on Wan2.1, from $0.2027$ to $0.1607$. Controlled comparisons with identical reconstruction rules and matched skip counts further confirm that these gains arise from allocating exact computation to more consequential denoising steps, rather than from using more computation or a stronger reconstructor. Beyond the main comparisons, the supplementary material provides an extensive measurement record, including full paired analyses with confidence intervals, a multiple-comparison audit, independent-seed replication, additional operating points, schedule-placement analysis, cost-model validation, and a codec-independence check.

Our contributions are summarized as follows:
\begin{itemize}
    \item We conduct controlled studies of the two fundamental dimensions of training-free feature reuse. The results show that, within the examined design space, more flexible reconstruction rules do not consistently outperform a bounded first-order reconstructor and can even be unstable, while exact-step placement remains a major source of quality variation.

    \item We show that raw shallow-probe variation is not directly comparable across denoising steps because it conflates trajectory displacement with network response. We further demonstrate that normalizing this variation by the realized probe-input movement provides a more effective online signal for exact-compute allocation.

    \item We propose CacheScope, a lightweight drop-in scheduling rule that converts trajectory-aware sensitivity into budget-constrained online exact-compute decisions. Experiments on image and video diffusion transformers demonstrate a new speed--fidelity frontier, with up to $3.57\times$ acceleration and up to $70.1\%$ lower median LPIPS than the strongest probe-based baseline.
\end{itemize}

\section{Related Work}
\label{sec:related_work}

\begin{figure*}[t]
    \centering
    \includegraphics[width=\textwidth]{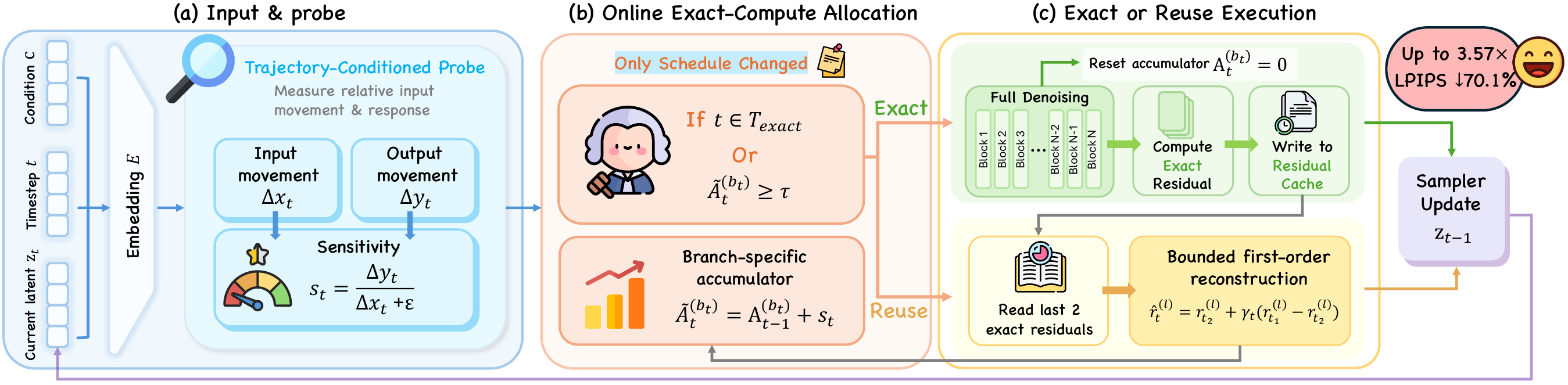}
     \vspace{-4mm}
    \caption{\textbf{Overview of CacheScope.} At every denoising step, the first
    transformer block is executed as a shallow probe. From its input and
    output we compute the relative trajectory movement $\Delta x_t$ and
    probe response $\Delta y_t$, and accumulate the trajectory-conditioned
    sensitivity $s_t=\Delta y_t/(\Delta x_t+\epsilon)$ --- the only change
    relative to raw probe scheduling. When the accumulator reaches the
    threshold $\tau$, the remaining blocks are evaluated exactly and the
    cache is refreshed; otherwise the step is reconstructed by the
    unchanged bounded first-order reconstructor. Under classifier-free
    guidance, the two branches maintain independent accumulators and
    cache states.}
    \label{fig:method_overview}
    \vspace{-5mm}
\end{figure*}

\subsection{Efficient Diffusion Inference}

Diffusion acceleration methods reduce either the number or the cost of
denoising evaluations~\cite{zhang2025drdiff}. Fast samplers use improved numerical solvers to
generate samples with fewer steps~\cite{ddim,dpmsolver,dpmsolverpp,
unipc,edm}; distillation compresses multi-step generation into a
single-step or few-step model~\cite{progressive_distillation,
consistency_models,lcm}; and quantization lowers the precision of
weights and activations~\cite{qdiffusion}; adjacent work on multimodal models~\cite{yao2025depthssc,zhang2025cfvlm,zhang2025mmcot,zhang2025srmcr} similarly reduces cost by token-level compression~\cite{hybridtokenvlm,zhang2026videoencoders}, and broader visual reasoning or coordination settings~\cite{zhang2025tdsr,zhang2025gamagent,zhang2025tri} offer related perspectives. Training-free feature reuse
is complementary to these directions: it preserves the pretrained
model and sampling trajectory while replacing selected full-network
evaluations with cached approximations.

\subsection{Training-Free Feature Reuse}

Feature caching exploits redundancy across adjacent denoising steps.
DeepCache reuses slowly varying U-Net features~\cite{deepcache}, while
FORA and $\Delta$-DiT extend fixed-stride reuse to diffusion
transformers~\cite{fora,delta_dit}. Later methods operate at different
granularities, including token-level reuse~\cite{toca}, attention
broadcasting~\cite{pab}, cross-branch reuse~\cite{fastercache,cfg},
and sample-adaptive intervals~\cite{adacache}.

Another line improves \emph{how to reuse}. TeaCache and MagCache model
feature or residual evolution to control reuse~\cite{teacache,
magcache}; TaylorSeer forecasts higher-order feature dynamics through
Taylor expansion~\cite{taylorseer}; and EasyCache reuses transformation
vectors according to online stability estimates~\cite{easycache}.
Our controlled experiments show that, once a stable bounded
first-order reconstructor is adopted, increasing reconstruction
complexity brings limited and inconsistent gains. We therefore keep
the reconstructor fixed and focus on \emph{when to reuse}.

\subsection{Adaptive Cache Scheduling}

Adaptive scheduling determines which steps should reuse cached features
and which should retain exact computation. Existing methods use
input-side changes, motion cues, transformation stability, or attention
statistics to adjust reuse intervals~\cite{teacache,adacache,easycache,
pab,tgate}.

DiCache is the closest work to ours. It runs a shallow probe online and
accumulates its absolute feature variation to trigger exact computation;
it also aligns multiple cached residuals for reconstruction~\cite{dicache}.
CacheScope retains the same probe-based setting and reconstructor, but
revisits the scheduling signal. We observe that absolute probe variation
conflates input-trajectory displacement with network response strength.
CacheScope removes this confounding by normalizing probe variation with
the corresponding input displacement, yielding an online directional
gain that better identifies risky reuse steps.

Related methods differ in how sensitivity or schedules are obtained.
SenCache also argues that sensitivity, rather than raw output change,
should drive caching decisions: it estimates finite-difference Jacobian
norms of the full model on a small offline calibration set and looks up
the resulting per-timestep sensitivity profiles at
inference~\cite{sencache}.CacheScope instead measures sensitivity \emph{online} from a shallow
probe along the realized denoising trajectory of the current sample.
It requires no offline sensitivity profiling or per-timestep lookup
table; only the scalar threshold $\tau$ is calibrated on a small
development set to match the target computation budget. At inference,
CacheScope uses a per-step response-per-movement ratio rather than a
precomputed sensitivity profile shared across samples.DPCache constructs an offline global schedule
through path-cost optimization, while CacheScope makes prompt-specific
decisions during inference~\cite{dpcache}. Concurrent ReCache learns a
budget-conditioned scheduling policy, whereas CacheScope requires no
policy training or auxiliary network~\cite{recache}. Thus, our
contribution is not a new reconstruction mechanism, but the
identification and removal of trajectory-motion confounding in online
probe-based scheduling.

\section{Method}
\label{sec:method}

CacheScope addresses a single question: given a base caching method,
at which of its reuse steps should a small budget of additional exact
evaluations be spent?
The resulting pipeline (Figure~\ref{fig:method_overview}) repeats a
five-stage loop: follow the fixed reuse policy, run a shallow probe,
form the output-over-input movement ratio, accumulate sensitivity
across reuse steps, and re-anchor with an exact evaluation once the
accumulated signal crosses a threshold.
CacheScope does not redesign how cached outputs are reconstructed; it
estimates how quickly the current cache anchor loses validity and
spends a small exact-computation budget to refresh it.

\subsection{Problem Formulation}
\label{sec:problem_formulation}

Consider a generation trajectory with $N$ denoiser evaluations. Let
$\mathbf z_t$ denote the latent before the $t$-th evaluation, $c$ the
condition, and $F_\theta$ the pretrained denoiser, whose output
$\mathbf r_t=F_\theta(\mathbf z_t,t,c)$ is consumed by the sampler
update $\Psi_t$.
A fixed \emph{base cache policy} $\{b_t\}_{t=1}^{N}$,
$b_t\in\{0,1\}$, specifies which steps the base method computes
exactly ($b_t{=}1$) and which it reuses from cache ($b_t{=}0$); it may
contain warm-up steps or any predefined reuse pattern and remains
unchanged throughout.
CacheScope introduces a binary \emph{fallback} $a_t$ only at nominal
reuse steps, giving the execution decision
\begin{equation}
    e_t
    =
    b_t \lor a_t,
    \qquad
    a_t \leq 1-b_t.
    \label{eq:final_execution}
\end{equation}
The output used by the sampler is
\begin{equation}
    \widetilde{\mathbf r}_t
    =
    \begin{cases}
        F_\theta(\mathbf z_t,t,c),
        & e_t=1,\\[1mm]
        \mathcal R_t(\mathcal H_{t-1}),
        & e_t=0,
    \end{cases}
    \label{eq:exact_or_cached}
\end{equation}
where $\mathcal R_t$ is the \emph{fixed} cache reconstructor and
$\mathcal H_{t-1}$ holds the exact states available before step $t$.
Writing $G_{\mathbf b,\mathbf a}$ for the final generation under
fallback decisions $\mathbf a$, $\mathbf a_\pi$ for the decisions
produced by an online policy $\pi$, and $G_{\mathrm{full}}$ for the
full-computation reference, the allocation problem is
\begin{equation}
\begin{aligned}
    \min_{\pi}\quad
    &\mathbb E\!\left[
        \mathcal D\!\left(
            G_{\mathbf b,\mathbf a_\pi},
            G_{\mathrm{full}}
        \right)
    \right] \\
    \mathrm{s.t.}\quad
    &\mathbb E\!\left[
        \sum_{t=1}^{N}(1-b_t)a_t
    \right]
    \leq B_{\mathrm{add}} .
\end{aligned}
\label{eq:allocation_objective}
\end{equation}
where $\pi$ is an online policy and $B_{\mathrm{add}}$ the budget of
additional exact evaluations.
Both the base schedule and the reconstructor are fixed; the only free
variable is \emph{where} the additional exact evaluations are placed.
This restriction separates CacheScope from methods that regenerate a
complete adaptive cache schedule and lets any improvement be
attributed to exact-compute placement alone.
Because the distortion $\mathcal D$ is unavailable online, CacheScope
instead builds an inexpensive signal indicating when the current cache
anchor becomes unreliable.

\subsection{Trajectory-Normalized Sensitivity}
\label{sec:normalized_sensitivity}

We split the denoiser as
$F_\theta=H_\theta\circ B_1\circ E_\theta$ and execute only the
prefix $B_1\circ E_\theta$ as a shallow probe:
\begin{equation}
    \mathbf x_t
    =
    E_\theta(\mathbf z_t,t,c),
    \qquad
    \mathbf y_t
    =
    B_1(\mathbf x_t),
    \label{eq:probe_decomposition}
\end{equation}
where $E_\theta$ is the input transformation, $B_1$ the first
transformer block, and $H_\theta$ the remaining blocks.
With $d(\cdot,\cdot)$ denoting the scale-normalized relative change
between consecutive states (mean absolute difference divided by the
$\epsilon$-stabilized mean magnitude of the earlier state;
Appendix~\ref{app:sensitivity_analysis}), the probe movements are
\begin{equation}
    \Delta x_t
    =
    d(\mathbf x_t,\mathbf x_{t-1}),
    \qquad
    \Delta y_t
    =
    d(\mathbf y_t,\mathbf y_{t-1}),
    \label{eq:input_output_movement}
\end{equation}
and the CacheScope signal is
\begin{equation}
    s_t
    =
    \frac{
        \Delta y_t
    }{
        \Delta x_t+\epsilon
    }.
    \label{eq:normalized_sensitivity}
\end{equation}

This normalization matters because raw probe variation conflates two
distinct factors. Let
$\mathbf v_t=\mathbf x_t-\mathbf x_{t-1}$ be the local trajectory
direction. A first-order expansion of the probe gives
\vspace{-1mm}
\begin{equation}
    \mathbf y_t-\mathbf y_{t-1}
    \approx
    J_t\mathbf v_t,
    \qquad
    \lambda_t
    =
    \frac{\left\|J_t\mathbf v_t\right\|}{\left\|\mathbf v_t\right\|},
    \label{eq:probe_linearization}
\end{equation}
with $J_t$ the local Jacobian of $B_1$, so that
\vspace{-1mm}
\begin{equation}
    \Delta y_t
    \approx
    \underbrace{\Delta x_t}_{\text{trajectory speed}}
    \;
    \underbrace{\lambda_t}_{\text{network sensitivity}},
    \qquad
    \frac{\Delta y_t}{\Delta x_t}
    \approx
    \lambda_t.
    \label{eq:product_decomposition}
\end{equation}
up to a slowly varying input--output scale ratio made precise in
Appendix~\ref{app:sensitivity_analysis}.
Raw probe variation therefore conflates trajectory speed with network
responsiveness; dividing out the former isolates the local directional
sensitivity $\lambda_t$.
Under a tight exact-computation budget, removing this nuisance factor
can shift the few threshold crossings that determine where anchors are
placed, and each shifted anchor changes every subsequent reuse step.
The full derivation, the resulting ranking-inversion condition, and a
detailed pathwise error analysis are provided in
Appendix~\ref{app:sensitivity_analysis}.

\subsection{Online Exact-Compute Allocation}
\label{sec:online_allocation}

A single $s_t$ measures one transition, but the same cache anchor
typically serves several consecutive reuse steps, and transitions that
are individually moderate can jointly invalidate it.
CacheScope therefore accumulates the sensitivity since the latest
exact evaluation and triggers a fallback once the accumulated value
crosses a calibrated threshold $\tau$:
\vspace{-1.5mm}
\begin{equation}
\begin{aligned}
    \widetilde A_t
    =
    A_{t-1}+s_t&,
    \qquad
    a_t
    =
    \mathbb I\!\left[
        b_t{=}0
        \land
        \widetilde A_t\geq\tau
    \right], \\
    A_t&
    =
    (1-e_t)\,\widetilde A_t .
\end{aligned}
\label{eq:allocation_rule}
\end{equation}
\vspace{-1.5mm}

Equation~\eqref{eq:allocation_rule} defines the online allocation rule
for one guidance branch. The accumulator $A_t$ tracks the sensitivity
accumulated since the latest exact evaluation, allowing either one
highly sensitive transition or several moderate transitions to trigger
a fallback. After an exact re-anchoring, $A_t$ is reset so that
sensitivity associated with the previous cache anchor is not carried
over to the refreshed state. Under local linearization, the accumulated
score serves as a lightweight proxy for the amplification of cache error
after the latest exact anchor
(Appendix~\ref{app:sensitivity_analysis}). The threshold $\tau$ is
calibrated once on a development set to match the target additional
exact-computation budget $B_{\mathrm{add}}$, without optimizing
test-set quality, and is then frozen. Under classifier-free guidance,
the two branches maintain independent probe states, accumulators, and
cache states (Appendix~\ref{app:cachescope_impl}).

\subsection{Re-Anchoring and Cost}
\label{sec:reanchoring_cost}

When a fallback triggers, the probe output is already available, so
the exact evaluation simply continues through the remaining blocks:
\begin{equation}
    \mathbf r_t
    =
    H_\theta(\mathbf y_t),
    \label{eq:resume_from_probe}
\end{equation}
after which the residual cache is refreshed and the accumulator is
reset.
At reuse steps, cached outputs are still produced by the base
configuration's bounded first-order reconstructor
$\mathcal R_t$~\cite{dicache}, which CacheScope leaves unchanged
(Appendix~\ref{app:cachescope_impl}).
An exact fallback therefore not only corrects the current output but
also replaces the cache anchor used by subsequent reuse steps; this
persistent downstream effect is why exact-compute allocation cannot be
reduced to independent per-step error minimization.
The overhead is minimal: at each reuse step CacheScope adds one
shallow block followed only by scalar reductions, and when a fallback
triggers, the exact evaluation resumes from the existing probe output,
so the probe is never recomputed
(Appendix~\ref{app:cachescope_impl}).

\section{Experiments}
\label{sec:experiments}

\begin{table*}[t]
\centering
\caption{\textbf{Main results.}
$\rho$ denotes the fraction of skipped model evaluations, and
Param.\ denotes each method's original speed-control hyperparameter.
Within each backbone and regime, the best and second-best fidelity
results in the target speed range are bolded and underlined,
respectively; ``--'' indicates that the target operating point is
unavailable.}
\label{tab:main_results}
\vspace{-1mm}
\small
\setlength{\tabcolsep}{2pt}
\renewcommand{\arraystretch}{0.9}
\begin{tabularx}{\linewidth}{@{}>{\raggedright\arraybackslash}p{1.9cm}
>{\centering\arraybackslash}p{1.4cm}
>{\raggedright\arraybackslash}p{3.8cm}|*{2}{>{\centering\arraybackslash}X}|*{2}{>{\centering\arraybackslash}X}|*{3}{>{\centering\arraybackslash}X}@{}}
\toprule
Backbone & Regime & Method & Param. & $\rho$ (\%)
& Lat. (s)$\downarrow$ & Speedup$\uparrow$
& LPIPS$\downarrow$ & PSNR$\uparrow$ & SSIM$\uparrow$ \\
\midrule

\multirow{10}{*}{
\shortstack[l]{
Wan2.1\\[-1pt]
{\tiny 50 steps, 81 frames}\\[-1pt]
{\tiny $832{\times}480$}
}}
& \multirow{5}{*}{Moderate}
& TeaCache~\cite{teacache} & $\delta{=}0.14$ & 60
& 102.41 & 2.40$\times$
& 0.1153 & 23.83 & 0.8676 \\
&
& MagCache~\cite{magcache} & $\delta{=}0.12$ & 58
& 107.10 & 2.29$\times$
& 0.0663 & 27.26 & 0.9153 \\
&
& EasyCache~\cite{easycache} & $\tau{=}0.058$ & 64
& 92.59 & 2.65$\times$
& 0.0909 & 24.89 & 0.8319 \\
&
& DiCache~\cite{dicache} & $\delta{=}0.16$ & 64
& 98.47 & 2.49$\times$
& \underline{0.0835} & \underline{25.78} & \underline{0.8958} \\
\cmidrule(lr){3-10}
&
& \cellcolor{gray!18}{\textbf{CacheScope}} & \cellcolor{gray!18}$\tau{=}5.5$ & \cellcolor{gray!18}64
& \cellcolor{gray!18}98.39 & \cellcolor{gray!18}2.49$\times$
& \cellcolor{gray!18}\textbf{0.0414} & \cellcolor{gray!18}\textbf{28.89} & \cellcolor{gray!18}\textbf{0.9220} \\
\cmidrule(lr){2-10}

&
\multirow{5}{*}{Fast}
& TeaCache~\cite{teacache} & $\delta{=}0.20$ & 72
& 73.76 & 3.33$\times$
& 0.2335 & 20.12 & 0.7358 \\
&
& MagCache~\cite{magcache} & -- & --
& -- & --
& -- & -- & -- \\
&
& EasyCache~\cite{easycache} & $\tau{=}0.20$ & 73
& 71.12 & 3.46$\times$
& 0.2761 & 20.40 & 0.6245 \\
&
& DiCache~\cite{dicache} & $\delta{=}0.50$ & 76
& 70.84 & 3.46$\times$
& \underline{0.2027} & \underline{20.87} & \textbf{0.7492} \\
\cmidrule(lr){3-10}
&
& \cellcolor{gray!18}\textbf{CacheScope} & \cellcolor{gray!18}$\tau{=}20.0$ & \cellcolor{gray!18}76
& \cellcolor{gray!18}70.82 & \cellcolor{gray!18}3.46$\times$
& \cellcolor{gray!18}\textbf{0.1607} & \cellcolor{gray!18}\textbf{21.80} & \cellcolor{gray!18}\underline{0.7363} \\

\midrule

\multirow{9}{*}{
\shortstack[l]{
Open-Sora 1.2\\[-1pt]
{\tiny 30 steps, 51 frames}\\[-1pt]
{\tiny 480p}
}}
& \multirow{5}{*}{Moderate}
& TeaCache~\cite{teacache} & $\delta{=}0.08$ & 27
& 41.97 & 1.36$\times$
& \underline{0.0421} & \underline{28.88} & \underline{0.9273} \\
&
& MagCache~\cite{magcache} & $\delta{=}0.003$ & 27
& 41.92 & 1.36$\times$
& 0.0732 & 25.37 & 0.8945 \\
&
& PAB~\cite{pab} & $R{=}2$ & 0
& 42.23 & 1.35$\times$
& 0.0783 & 26.92 & 0.8864 \\
&
& DiCache~\cite{dicache} & $\delta{=}0.03$ & 30
& 40.69 & 1.40$\times$
& 0.0566 & 26.27 & 0.9141 \\
\cmidrule(lr){3-10}
&
& \cellcolor{gray!18}\textbf{CacheScope} & \cellcolor{gray!18}$\tau{=}2.75$ & \cellcolor{gray!18}30
& \cellcolor{gray!18}40.69 & \cellcolor{gray!18}1.40$\times$
& \cellcolor{gray!18}\textbf{0.0169} & \cellcolor{gray!18}\textbf{34.95} & \cellcolor{gray!18}\textbf{0.9623} \\
\cmidrule(lr){2-10}

&
\multirow{4}{*}{Fast}
& TeaCache~\cite{teacache} & $\delta{=}0.17$ & 53
& 26.81 & 2.13$\times$
& 0.1740 & 21.53 & 0.8122 \\
&
& MagCache~\cite{magcache} & $\delta{=}0.01766$ & 53
& 26.73 & 2.14$\times$
& \underline{0.1242} & \underline{23.03} & \underline{0.8344} \\
&
& DiCache~\cite{dicache} & $\delta{=}0.07$ & 53
& 27.87 & 2.05$\times$
& 0.1293 & 22.07 & 0.8294 \\
\cmidrule(lr){3-10}
&
& \cellcolor{gray!18}\textbf{CacheScope} & \cellcolor{gray!18}$\tau{=}6.0$ & \cellcolor{gray!18}53
& \cellcolor{gray!18}27.83 & \cellcolor{gray!18}2.05$\times$
& \cellcolor{gray!18}\textbf{0.0664} & \cellcolor{gray!18}\textbf{26.13} & \cellcolor{gray!18}\textbf{0.9004} \\

\midrule

\multirow{12}{*}{
\shortstack[l]{
FLUX.1-dev\\[-1pt]
{\tiny 50 steps}\\[-1pt]
{\tiny $1024{\times}1024$}
}}
& \multirow{6}{*}{Moderate}
& TeaCache~\cite{teacache} & $\delta{=}0.55$ & 66
& 10.47 & 2.899$\times$
& 0.2843 & 18.47 & 0.7730 \\
&
& MagCache~\cite{magcache} & $\delta{=}0.17$ & 66
& 10.49 & 2.893$\times$
& \underline{0.1209} & 24.88 & 0.8747 \\
&
& EasyCache~\cite{easycache} & $\tau{=}0.18$ & 66
& 10.48 & 2.897$\times$
& 0.1680 & \underline{25.93} & \underline{0.8786} \\
&
& TaylorSeer~\cite{taylorseer} & $N{=}3$ & 64
& 11.82 & 2.568$\times$
& 0.2045 & 19.90 & 0.8095 \\
&
& DiCache~\cite{dicache} & $\delta{=}0.15$ & 68
& 10.27 & 2.954$\times$
& 0.1353 & 23.32 & 0.8769 \\
\cmidrule(lr){3-10}
&
& \cellcolor{gray!18}\textbf{CacheScope} & \cellcolor{gray!18}$\tau{=}4.6$ & \cellcolor{gray!18}68
& \cellcolor{gray!18}10.28 & \cellcolor{gray!18}2.954$\times$
& \cellcolor{gray!18}\textbf{0.0816} & \cellcolor{gray!18}\textbf{27.74} & \cellcolor{gray!18}\textbf{0.9219} \\
\cmidrule(lr){2-10}

&
\multirow{6}{*}{Fast}
& TeaCache~\cite{teacache} & $\delta{=}0.71$ & 70
& 9.25 & 3.311$\times$
& 0.3268 & 16.87 & 0.7303 \\
&
& MagCache~\cite{magcache} & $\delta{=}0.48$ & 72
& 8.67 & 3.499$\times$
& 0.1913 & 21.97 & 0.8300 \\
&
& EasyCache~\cite{easycache} & $\tau{=}0.36$ & 72
& 8.67 & 3.503$\times$
& 0.2978 & \underline{23.88} & 0.8083 \\
&
& TaylorSeer~\cite{taylorseer} & $N{=}5$ & 76
& 8.37 & 3.624$\times$
& 0.3157 & 17.11 & 0.7289 \\
&
& DiCache~\cite{dicache} & $\delta{=}0.25$ & 74
& 8.49 & 3.601$\times$
& \underline{0.1842} & 21.40 & \underline{0.8333} \\
\cmidrule(lr){3-10}
&
& \cellcolor{gray!18}\textbf{CacheScope} & \cellcolor{gray!18}$\tau{=}6.4$ & \cellcolor{gray!18}74
& \cellcolor{gray!18}8.51 & \cellcolor{gray!18}3.566$\times$
& \cellcolor{gray!18}\textbf{0.1246} & \cellcolor{gray!18}\textbf{24.99} & \cellcolor{gray!18}\textbf{0.8760} \\

\bottomrule
\end{tabularx}

\vspace{-3mm}
\end{table*}

\begin{table}[t]
    \centering
    \small
    \caption{
    \textbf{Controlled scheduler comparison on $32$ held-out prompts.}
    The three signals (D, M, S) share the same DiCache bounded
first-order $\Gamma$ reconstructor, probe depth, and warm-up policy,
and differ only in the accumulated scheduling signal.
    Values are medians over prompts.
    Statistical testing uses the strictly matched-skip subsets described
in Appendix~\ref{app:full_paired}.
    Best values within each backbone and regime are in bold;
    second-best values are underlined.
    }
    \label{tab:scheduler_comparison}
    \setlength{\tabcolsep}{5.5pt}
     \resizebox{\columnwidth}{!}{%
    \begin{tabular}{llccccc}
        \toprule
        Backbone
        & Regime
        & Scheduler
        & Speedup $\uparrow$
        & LPIPS $\downarrow$
        & PSNR $\uparrow$
        & SSIM $\uparrow$ \\
        \midrule
        
        \multirow{6}{*}{Wan2.1}
        & \multirow{3}{*}{Moderate}
        & D: $\Delta y$
        & 2.492
        & 0.0772
        & 26.010
        & 0.8961 \\
        
        &
        & M: $|\Delta y-\Delta x|$
        & 2.491
        & \underline{0.0470}
        & \underline{28.82}
        & \textbf{0.9329} \\
        
        &
        & \textbf{S: Probe-normalized}
        & 2.493
        & \textbf{0.0414}
        & \textbf{28.891}
        & \underline{0.9220} \\
        \cmidrule(lr){2-7}
        
        &
        \multirow{3}{*}{Fast}
        & D: $\Delta y$
        & 3.462
        & \underline{0.2120}
        & \underline{21.013}
        & \textbf{0.7437} \\
        
        &
        & M: $|\Delta y-\Delta x|$
        & 3.463
        & 0.2517
        & 19.69
        & 0.6131 \\
        
        &
        & \textbf{S: Probe-normalized}
        & 3.463
        & \textbf{0.1607}
        & \textbf{21.802}
        & \underline{0.7363} \\
        
        \midrule
        
        \multirow{6}{*}{FLUX}
        & \multirow{3}{*}{Moderate}
        & D: $\Delta y$
        & 2.954
        & 0.1353
        & 23.320
        & 0.8769 \\
        
        &
        & M: $|\Delta y-\Delta x|$
        & 2.953
        & \underline{0.1014}
        & \underline{26.177}
        & \underline{0.9035} \\
        
        &
        & \textbf{S: Probe-normalized}
        & 2.954
        & \textbf{0.0816}
        & \textbf{27.745}
        & \textbf{0.9219} \\
        \cmidrule(lr){2-7}
        
        &
        \multirow{3}{*}{Fast}
        & D: $\Delta y$
        & 3.601
        & 0.1842
        & 21.400
        & 0.8333 \\
        
        &
        & M: $|\Delta y-\Delta x|$
        & 3.568
        & \underline{0.1474}
        & \underline{24.358}
        & \underline{0.8626} \\
        
        &
        & \textbf{S: Probe-normalized}
        & 3.566
        & \textbf{0.1246}
        & \textbf{24.995}
        & \textbf{0.8760} \\
        
        \bottomrule
    \end{tabular}}
    \vspace{-5mm}
    
\end{table}

\subsection{Experimental Setup}
\label{sec:experimental_setup}

\paragraph{Setup.}
We evaluate on FLUX.1-dev~\cite{flux},
Wan2.1-T2V-1.3B~\cite{wan}, and Open-Sora
1.2~\cite{opensora}, using the generation protocols listed in
Table~\ref{tab:main_results}. For each backbone, scheduling thresholds
are calibrated on a 16-prompt development set and frozen before
evaluation on a disjoint 32-prompt test set, drawn from
DrawBench~\cite{imagen} for FLUX and VBench~\cite{vbench} for the video
models. Keeping the probe, reconstructor, cache policy, and model
implementation fixed, we compare the DiCache signal
$\Delta y_t$, the difference signal
$|\Delta y_t-\Delta x_t|$, and our normalized signal
$\Delta y_t/(\Delta x_t+\epsilon)$ under matched skip counts at two
acceleration regimes. We report LPIPS~\cite{lpips}, PSNR,
SSIM~\cite{ssim}, latency, and speedup, with paired prompt-level
comparisons. Full generation settings, hyperparameters, hardware,
random seeds, and statistical protocols are provided in the
supplementary material.

\subsection{Main Results}
\label{sec:main_results}

Table~\ref{tab:main_results} reports the median quality and measured
speedup over the $32$ held-out prompts, and three observations stand
out.
First, the improvement is consistent across all three backbones: at
matched speedups, CacheScope achieves the lowest median LPIPS in every
backbone--regime combination, on the image transformer FLUX.1-dev as
well as the video transformers Wan2.1 and Open-Sora 1.2.
Second, the largest gain appears on Open-Sora at the Moderate level, where
LPIPS drops from DiCache's $0.0566$ to $0.0169$ at identical
compute---a $70.1\%$ reduction.
Third, the advantage persists in the most aggressive regimes: at about
$3.57\times$ on FLUX and $3.46\times$ on Wan2.1, CacheScope still
lowers median LPIPS from $0.1842$ to $0.1246$ ($32.4\%$) and from
$0.2027$ to $0.1607$ ($20.7\%$) relative to the strongest baseline at
the same operating point, so the gains are not confined to
conservative caching settings.
Baseline hyperparameters are selected on the development set to place
each method inside the target speed range of its regime
(Appendix~\ref{app:baseline_params}).

\begin{figure*}[t]
    \centering
    \includegraphics[width=\textwidth]{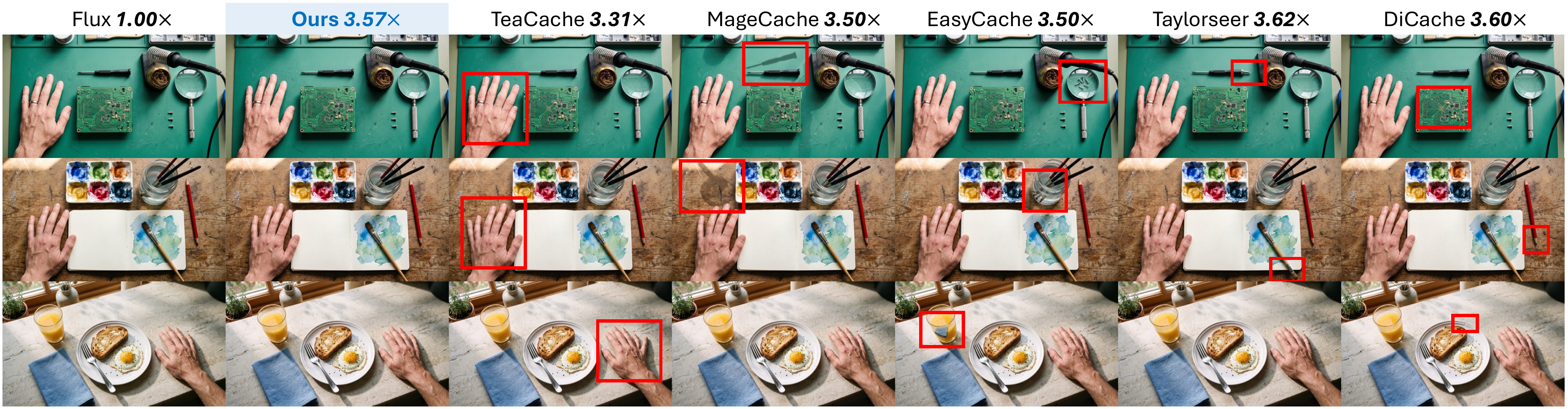}
    \vspace{-6mm}
    \caption{
    \textbf{Qualitative comparison on FLUX.1-dev in the Fast regime.}
    Columns show the dense output and different training-free caching
    methods, while rows correspond to three prompts. Red boxes highlight
    visible deviations from the dense reference. CacheScope better
    preserves fine structures and object geometry at $3.57\times$ speedup.
    }
    \label{fig:qualitative}
    \vspace{-4.5mm}
\end{figure*}

Table~\ref{tab:scheduler_comparison} isolates the effect of the
scheduling signal on the two backbones for which all three signals
were run under the controlled protocol, Wan2.1 and FLUX, by holding
the reconstructor, probe depth, and warm-up policy fixed.
Relative to the default DiCache signal D, probe normalization reduces
median LPIPS by $46.4\%$ and $24.2\%$ on Wan2.1 and by $39.7\%$ and
$32.4\%$ on FLUX in the Moderate and Fast regimes, respectively;
relative to the stronger difference-based signal M, the reductions
range from $11.9\%$ to $36.2\%$.
Because the three schedulers execute the same shallow probe and nearly
identical numbers of exact forwards, their speed differences are
negligible, which isolates the effect of exact-step placement from the
amount of computation.

\paragraph{Statistical reliability.}
Paired prompt-level LPIPS differences
$\Delta L=L_{\mathrm{S}}-L_{\mathrm{baseline}}$ under strictly equal
skip counts, where negative values favor our scheduler, are reported in
full---with confidence intervals and win--tie--loss counts---in
Appendix~\ref{app:full_paired} (Table~\ref{tab:app_paired_full}).
Our scheduler significantly outperforms both comparison signals in all
eight LPIPS comparisons, and the four primary S-versus-D comparisons
remain significant after Benjamini--Hochberg correction.
The improvements replicate with an independent seed ($43$) at similar
effect magnitudes, and the conclusions do not reverse when all $32$
prompts are analyzed without matched-skip filtering; median skip-count
differences lie in $\{-1,0\}$, so the gains cannot be explained by
skipping fewer evaluations (full replication and robustness tables are
given in the supplementary material).
On FLUX, the LPIPS improvements are accompanied by consistent gains in
PSNR and SSIM; on Wan2.1, SSIM exhibits isolated trade-offs, so we
treat perceptual fidelity (LPIPS) as the primary outcome rather than
claiming uniform dominance on every metric.

\subsection{Ablation and Overhead}
\label{sec:ablation_overhead}

\begin{figure}[t]
    \centering
    \includegraphics[width=\columnwidth]{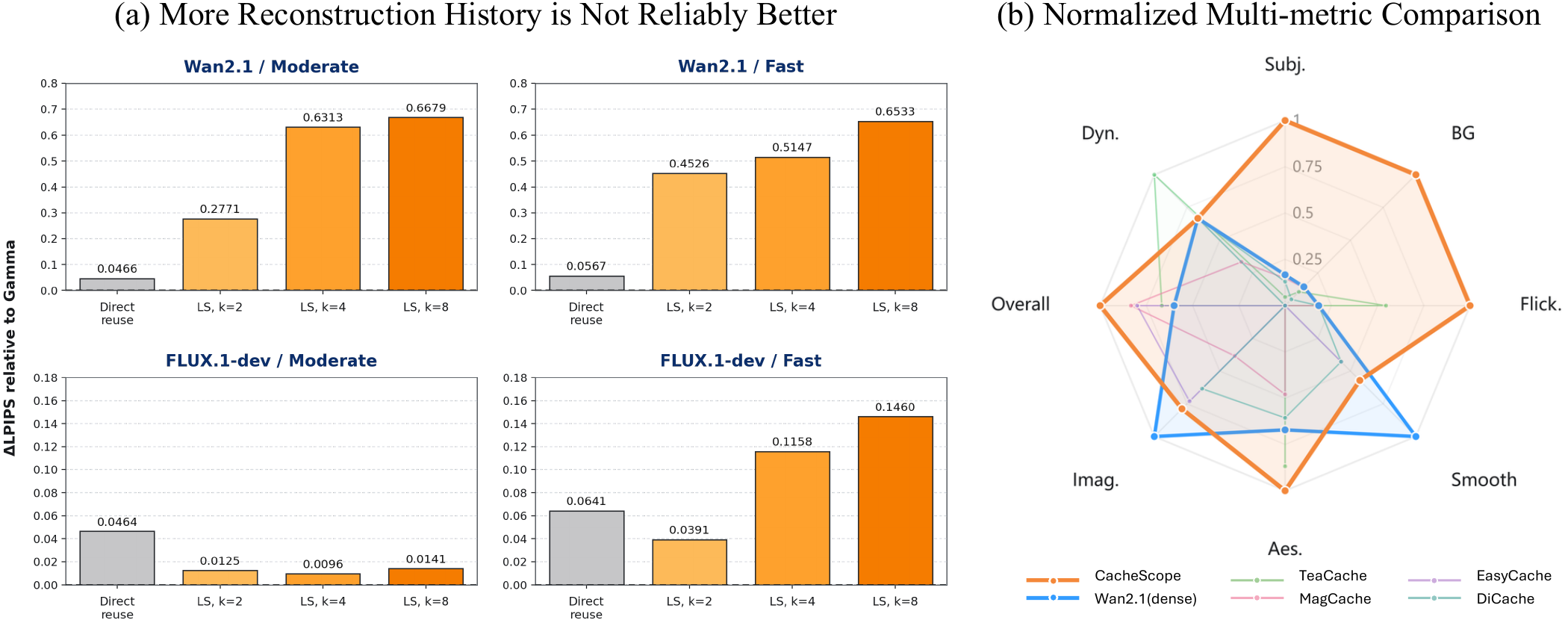}
     \vspace{-6mm}
    \caption{\textbf{Reconstruction ablation and multi-metric
    comparison.} (a) $\Delta$LPIPS relative to the bounded
    first-order $\Gamma$ reconstructor under identical replayed
    exact/skip schedules (positive $=$ worse). (b) VBench
    sub-dimension scores on Wan2.1 (Moderate), \emph{min--max
    normalized per metric across methods}; raw scores are given in
    Appendix~\ref{app:vbench_eval}.}
    \label{fig:recon_history}
    \vspace{-5mm}
\end{figure}

\paragraph{Reconstruction complexity.}
Figure~\ref{fig:recon_history}(a) examines whether further
increasing reconstruction complexity provides a comparable benefit; all
variants replay the same exact/skip trace, so the comparison changes
only the reconstruction rule (absolute median LPIPS and per-prompt
win--loss counts are given in
Appendix~\ref{app:reconstruction_extended}).

Across the four settings, none of the unconstrained multi-history
least-squares variants consistently improves over the bounded
first-order $\Gamma$ reconstructor. The degradation becomes pronounced
and increases with history length on Wan2.1, whereas the changes on
FLUX are smaller and non-monotonic. These results support the narrower
conclusion that, within the examined design space, additional
reconstruction flexibility is not a reliable source of gains. We
therefore fix a stable bounded first-order reconstructor and attribute
the primary improvements of CacheScope to exact-compute scheduling
rather than to a more flexible residual predictor.

\paragraph{Computational overhead.}
The additional scheduler computation is limited to evaluating
$\Delta x_t$ and one scalar division.
At matched operating points, the observed latency difference between S
and D is below $0.15\%$ on FLUX and $0.1\%$ on Wan2.1, and the measured
speedup reaches $0.90$--$0.99$ of the analytical FLOPs estimate, with
the remaining gap explained by fixed costs such as text encoding, VAE
decoding, and scheduler bookkeeping.
Probe normalization thus changes \emph{where} exact computation is
allocated without materially changing the total inference cost.

\section{Conclusion}

We revisited training-free feature reuse for diffusion transformers and
showed that, once a stable bounded first-order reconstructor is fixed,
the placement of exact evaluations is a dominant factor for generation
fidelity, while the more flexible reconstruction formulations we
examined did not consistently outperform the first-order rule. We
further identified that raw probe variation is not a directly comparable
measure of step importance, because it conflates diffusion-trajectory
displacement with the network response induced by that displacement.
Motivated by this, we proposed CacheScope, which schedules exact
computation from a trajectory-conditioned sensitivity score and converts
it into a budget-constrained online allocation policy without any
training, gradients, auxiliary model, or offline schedule search. Across
image and video diffusion transformers, CacheScope establishes a new
speed--fidelity frontier under matched exact-compute budgets. A current
limitation is that sensitivity is estimated from a single shallow probe
along the realized denoising direction; multi-probe and higher-order
trajectory signals remain future work.
\bibliography{main}
\clearpage
\appendix
\setcounter{secnumdepth}{2}
\setcounter{section}{0}
\renewcommand{\thesection}{\Alph{section}}
\def\CacheScopeMainFile{1}
\makeatletter
\@ifundefined{CacheScopeMainFile}{%
  \newif\ifSuppStandalone
  \SuppStandalonetrue
}{%
  \newif\ifSuppStandalone
  \SuppStandalonefalse
}
\makeatother

\ifSuppStandalone
\documentclass[letterpaper]{article}

\usepackage[submission]{aaai2027}
\usepackage[hyphens]{url}
\usepackage{graphicx}
\urlstyle{rm}
\def\UrlFont{\rm}
\usepackage{natbib}
\usepackage{caption}
\usepackage{booktabs}
\usepackage{amsmath}
\usepackage{amssymb}
\usepackage{multirow}
\usepackage{xcolor}
\usepackage{tabularx}
\usepackage{array}
\frenchspacing

\pdfinfo{
/TemplateVersion (2027.1)
}

\newcommand{\method}{CacheScope}
\newcommand{\shortmethod}{CacheScope}

\setcounter{secnumdepth}{2}
\setcounter{section}{0}
\renewcommand{\thesection}{\Alph{section}}
\renewcommand{\thesubsection}{\Alph{section}.\arabic{subsection}}
\renewcommand{\thetable}{\Alph{table}}
\renewcommand{\thefigure}{\Alph{figure}}
\renewcommand{\theequation}{\Alph{section}.\arabic{equation}}

\begin{document}
\fi

\twocolumn[
{\centering
{\LARGE\bfseries Supplementary Material\par}
\vspace{0.6em}
{\large CacheScope: Exact-Compute Allocation\par}
{\large for Training-Free Diffusion Acceleration\par}
\vspace{1em}
}
]
\section{Relationship to Existing Scheduling Signals}
\label{app:signal_comparison}

We consider three scheduling signals:
\begin{equation}
\sigma_t^{D}
=
\Delta y_t,
\qquad
\sigma_t^{M}
=
\left|\Delta y_t-\Delta x_t\right|,
\qquad
\sigma_t^{S}
=
\frac{\Delta y_t}{\Delta x_t+\epsilon}.
\label{eq:app_three_signals}
\end{equation}

Here, $\sigma_t^{D}$ is raw probe-output variation,
$\sigma_t^{M}$ is the difference-based alternative, and
$\sigma_t^{S}$ is the trajectory-conditioned sensitivity used by
CacheScope.

The difference-based and ratio-based signals encode distinct
decision semantics. When the probe response is much weaker than
the trajectory movement,
\begin{equation}
\Delta y_t \ll \Delta x_t,
\end{equation}
we have
\begin{equation}
\sigma_t^{M}
\approx
\Delta x_t,
\qquad
\sigma_t^{S}
\ll 1.
\label{eq:app_weak_response}
\end{equation}
The difference-based signal may therefore assign a large score to a
step whose network response is weak relative to its input movement.

When the input and output variations have similar magnitudes,
\begin{equation}
\Delta y_t
\approx
\Delta x_t,
\end{equation}
we instead have
\begin{equation}
\sigma_t^{M}
\approx
0,
\qquad
\sigma_t^{S}
\approx
1.
\label{eq:app_proportional_response}
\end{equation}

The two signals are not related by a global scalar transformation.
Ignoring $\epsilon$ for illustration, consider
\begin{equation}
(\Delta x,\Delta y)=(1,0.5)
\end{equation}
and
\begin{equation}
(\Delta x,\Delta y)=(2,1).
\end{equation}
Both cases produce
\begin{equation}
\sigma^{S}=0.5,
\end{equation}
but their difference-based scores are
\begin{equation}
\sigma^{M}=0.5
\qquad\text{and}\qquad
\sigma^{M}=1,
\end{equation}
respectively. Therefore, no globally defined scalar function $h$
can satisfy
\begin{equation}
\sigma_t^{M}
=
h(\sigma_t^{S})
\label{eq:app_no_mapping}
\end{equation}
for every valid pair $(\Delta x_t,\Delta y_t)$.

The two signals may also reverse the ranking of denoising steps.
For example,
\begin{equation}
(\Delta x_1,\Delta y_1)=(1,0.2),
\qquad
(\Delta x_2,\Delta y_2)=(0.2,0.1)
\end{equation}
give
\begin{equation}
\sigma_1^{M}=0.8
>
\sigma_2^{M}=0.1,
\end{equation}
but
\begin{equation}
\sigma_1^{S}=0.2
<
\sigma_2^{S}=0.5.
\end{equation}
Thus, difference-based scheduling prioritizes the first step,
whereas CacheScope assigns greater sensitivity to the second step.

\subsection{Offline Signal Aggregation and Schedule Information}

\paragraph{Motivation.}
We further examine whether the advantage of the
trajectory-normalized signal depends on prompt-specific online
fluctuations, or whether it also reveals a stable timestep structure
shared across prompts. We construct two fixed schedules using only the
16-prompt development set. For \textsc{Global-S}, we independently
average the normalized signal
\begin{equation}
    s_t
    =
    \frac{\Delta y_t}{\Delta x_t+\epsilon}
\end{equation}
at each timestep and apply the same accumulator-and-threshold rule used
by CacheScope to the resulting mean curve. For \textsc{Global-D}, we
repeat exactly the same procedure using the raw probe variation
$\Delta y_t$. The threshold for each schedule is selected solely to
match the target median skip count of 34 steps; no fidelity metric is
used during schedule construction.

Both fixed masks are generated before any validation output or
evaluation metric is accessed. The development, validation, and
intervention prompt sets are mutually disjoint. We additionally
evaluate cross-prompt replay, in which the \textsc{Online-S} schedule
generated for prompt $i$ is applied to prompt $i+1$ under a cyclic
derangement. All variants continue to execute the same shallow probe;
only their exact-versus-skip decisions are replaced. The comparison
therefore isolates the information contained in the scheduling signal
and its prompt-specific assignment. These fixed schedules are used
only for diagnostic analysis and are not part of the deployed
CacheScope inference procedure.

\begin{table*}[t]
\centering
\small
\setlength{\tabcolsep}{4.5pt}
\caption{
\textbf{Offline aggregation and schedule-information analysis on
FLUX.1-dev at the Moderate operating point.}
\textsc{Global-S} and \textsc{Global-D} are constructed from the same
16-prompt development set using the same aggregation procedure,
threshold solver, and target compute budget; they differ only in the
scheduling signal. Metric values are medians over 32 held-out prompts.
The exact-evaluation column reports totals across all 32 prompts.
All variants have a median skip count of 34. Paired two-sided sign
tests compare each variant with \textsc{Online-S}.
}
\label{tab:offline_schedule_information}
\resizebox{\textwidth}{!}{%
\begin{tabular}{llccrrrr}
\toprule
Scheduler
& Signal/source
& Prompt assignment
& Total exact evals
& LPIPS $\downarrow$
& PSNR $\uparrow$
& SSIM $\uparrow$
& Paired LPIPS $p$ \\
\midrule

\textsc{Online-S}
& $s_t$, online
& Matched
& 515
& 0.0816
& 27.74
& 0.9219
& --- \\

\textsc{Global-S}
& Mean $s_t$, dev16
& Global
& 512
& 0.0792
& 27.54
& 0.9202
& 1.00 \\

\textsc{Global-D}
& Mean $\Delta y_t$, dev16
& Global
& 512
& 0.1337
& 23.31
& 0.8734
& $2.6{\times}10^{-6}$ \\

Cross-prompt replay
& Mismatched \textsc{Online-S} masks
& Deranged
& 515
& 0.0794
& 27.35
& 0.9201
& 1.00 \\

\bottomrule
\end{tabular}%
}
\vspace{-3mm}
\end{table*}

\paragraph{Results.}
We detect no significant difference between \textsc{Global-S} and
\textsc{Online-S}. The median LPIPS of \textsc{Global-S} is numerically
lower, at 0.0792 compared with 0.0816 for \textsc{Online-S}, with 14
improvements, 13 degradations, and five ties across the 32 paired
prompts ($p=1.00$). Cross-prompt replay likewise shows no significant
difference from \textsc{Online-S}. At this FLUX operating point, these
results suggest that a substantial portion of the useful scheduling
structure is shared across prompts, while prompt-specific schedule
variation provides no additional statistically detectable benefit.

Importantly, the same offline construction does not recover this
shared structure when applied to the raw DiCache signal. Compared with
\textsc{Global-S}, \textsc{Global-D} increases median LPIPS from
0.0792 to 0.1337; relative to \textsc{Online-S}, it is worse on 29 of
the 32 paired prompts ($p=2.6{\times}10^{-6}$).
\textsc{Global-S} and \textsc{Global-D} use identical development
data, schedule-construction procedures, and compute budgets, and
differ only in whether the probe variation is normalized by the
corresponding input movement. This controlled comparison indicates
that trajectory normalization factors out variation attributable to
input movement and exposes a substantially more effective
timestep-importance profile for allocating exact computation.

\paragraph{Schedule analysis.}
The online schedules are not identical: the 32 prompts produce nine
distinct masks, with a median pairwise Jaccard similarity of 0.200
within the adaptive scheduling window. The schedule-dependent
exact-step sets of \textsc{Global-S} and \textsc{Global-D} within this
window also differ substantially, with a Jaccard similarity of 0.286.
Their respective locations are
$\{7,9,12,16,19,25,32,38,44\}$ and
$\{9,14,19,26,32,37,41,44,47\}$.
All remaining mandatory exact evaluations are shared between the two
schedules. Thus, their large fidelity difference corresponds to a
meaningful change in exact-compute placement rather than to two nearly
identical masks.

\section{Detailed Analysis of the Sensitivity Signal}
\label{app:sensitivity_analysis}

This appendix provides the full derivations behind the
trajectory-normalized sensitivity of the main text: the precise
definition of the relative-change measure, the scale-normalized form
of the signal, the ranking-inversion condition under raw probe
variation, and a pathwise error analysis of the accumulated score.

\subsection{Relative-Change Measure and Scale-Normalized Form}
\label{app:scale_normalized}

For tensors $\mathbf u$ and $\mathbf v$, the scale-normalized movement
used throughout is
\begin{equation}
    d(\mathbf u,\mathbf v)
    =
    \frac{
        \mu\!\left(
            \left|\mathbf u-\mathbf v\right|
        \right)
    }{
        \mu\!\left(
            \left|\mathbf v\right|
        \right)+\epsilon
    },
    \label{eq:relative_movement}
\end{equation}
where $\mu(\cdot)$ denotes the mean over all non-batch dimensions and
$\epsilon=10^{-6}$ is a numerical constant.

Substituting the first-order expansion
$\mathbf y_t-\mathbf y_{t-1}\approx J_t\mathbf v_t$ into the
implemented relative measures shows that the score approximates
\begin{equation}
    s_t
    \propto
    \frac{
        \left\|J_t\mathbf v_t\right\|
    }{
        \left\|\mathbf v_t\right\|
    }
    \cdot
    \frac{
        \left\|\mathbf x_{t-1}\right\|
    }{
        \left\|\mathbf y_{t-1}\right\|
    }.
    \label{eq:scale_normalized_gain}
\end{equation}
The first factor is the directional response gain $\lambda_t$ of the
main text; the second provides relative scale normalization between
the probe input and output. Hence $s_t$ can be interpreted as a
scale-normalized directional response gain of the shallow denoiser.
It measures the model response only along the direction actually
traversed by the trajectory, which is the direction relevant for cache
scheduling; it is not an estimate of the full Jacobian norm.

\subsection{Ranking Inversion under Raw Probe Variation}
\label{app:ranking_inversion}

The product decomposition
$\Delta y_t\approx\Delta x_t\lambda_t$ implies that raw probe
variation can invert the ordering induced by sensitivity. Consider two
candidate reuse steps $i$ and $j$ with
\begin{equation}
    \lambda_i < \lambda_j.
    \label{eq:true_sensitivity_order}
\end{equation}
Although step $j$ is more sensitive, the raw variations satisfy
\begin{equation}
    \Delta y_i
    >
    \Delta y_j
    \quad
    \text{if}
    \quad
    \frac{
        \Delta x_i
    }{
        \Delta x_j
    }
    >
    \frac{
        \lambda_j
    }{
        \lambda_i
    },
    \label{eq:ranking_inversion}
\end{equation}
in which case ranking or thresholding the raw signal prioritizes the
\emph{less} sensitive step because of its larger trajectory
displacement. Normalization removes this nuisance factor and restores
the ordering induced by the directional response gain.
When only a small fraction of reuse steps can be replaced by exact
evaluations, allocation quality is dominated by the few steps close to
the decision boundary; correcting even a limited number of such
ranking inversions can considerably change the selected re-anchoring
positions.

\subsection{Pathwise Error Analysis}
\label{app:pathwise_error}

The accumulated score admits a pathwise error interpretation. An exact
evaluation on an already approximate latent does not recover the
original full trajectory; it does, however, remove the additional
cache approximation error relative to the current trajectory state.
Accordingly, we analyze the error introduced \emph{after} the latest
exact re-anchoring.

Suppose the latest exact evaluation occurs at step $t_a$. Define a
counterfactual reference trajectory
$\{\overline{\mathbf z}_t\}_{t\geq t_a}$ that starts from the current
approximate latent,
$\overline{\mathbf z}_{t_a}=\widetilde{\mathbf z}_{t_a}$, and uses exact
denoiser evaluations thereafter:
\begin{equation}
    \overline{\mathbf z}_{t+1}
    =
    \Psi_t
    \left(
        \overline{\mathbf z}_t,
        F_\theta(\overline{\mathbf z}_t,t,c)
    \right).
    \label{eq:anchored_reference_trajectory}
\end{equation}
Let
$\boldsymbol{\delta}_t=\widetilde{\mathbf z}_t-\overline{\mathbf z}_t$
denote the additional trajectory deviation introduced by caching after
the latest re-anchoring (so $\boldsymbol{\delta}_{t_a}=\mathbf 0$), and
let
$\boldsymbol{\varepsilon}_t
=\widetilde{\mathbf r}_t-F_\theta(\widetilde{\mathbf z}_t,t,c)$
denote the cache reconstruction error relative to an exact evaluation
at the same latent. A local linearization of the sampler update gives
\begin{equation}
    \boldsymbol{\delta}_{t+1}
    \approx
    M_t\boldsymbol{\delta}_t
    +
    N_t\boldsymbol{\varepsilon}_t,
    \label{eq:error_dynamics}
\end{equation}
where $M_t$ and $N_t$ are the local Jacobians of the sampler update
with respect to its latent and denoiser-output arguments.

To connect probe sensitivity with cache mismatch, we adopt a local
probe-consistency assumption:
\begin{equation}
    \left\|
        \boldsymbol{\varepsilon}_t
    \right\|
    \leq
    \eta_t
    +
    \beta_t s_t
    \left\|
        \boldsymbol{\delta}_t
    \right\|,
    \label{eq:probe_consistency}
\end{equation}
where $\eta_t$ is the intrinsic error of the fixed reconstructor and
$\beta_t$ a bounded local coefficient: an existing trajectory
deviation is more likely to produce cache mismatch when the denoiser
is locally sensitive along the current direction.
Combining Eqs.~\eqref{eq:error_dynamics} and
\eqref{eq:probe_consistency} yields
\begin{equation}
    \left\|
        \boldsymbol{\delta}_{t+1}
    \right\|
    \leq
    \left(
        m_t+c_t s_t
    \right)
    \left\|
        \boldsymbol{\delta}_t
    \right\|
    +
    \xi_t,
    \label{eq:one_step_error_bound}
\end{equation}
with $m_t=\|M_t\|$, $c_t=\|N_t\|\beta_t$, and $\xi_t=\|N_t\|\eta_t$.
Writing $\overline c_t=c_t/m_t$ and unrolling from the latest anchor,
then applying $1+u\leq\exp(u)$ for $u\geq0$, gives
\begin{equation}
\begin{aligned}
    \left\|
        \boldsymbol{\delta}_t
    \right\|
    \leq
    \sum_{j=t_a}^{t-1}
    \xi_j
    \left(
        \prod_{k=j+1}^{t-1}
        m_k
    \right)
    \exp
    \left(
        \sum_{k=j+1}^{t-1}
        \overline c_k s_k
    \right).
\end{aligned}
\label{eq:error_exponential_bound}
\end{equation}
The factors $\prod_k m_k$ describe the underlying sampler dynamics and
are shared by all allocation signals; the sensitivity-dependent
amplification is controlled by the cumulative quantity
$\sum_k\overline c_k s_k$. When the bounded coefficients
$\overline c_k$ vary slowly or are absorbed into the threshold, the
unweighted accumulator $A_t=\sum_{k=t_a+1}^{t}s_k$ serves as a
lightweight surrogate for this amplification, which motivates
accumulating $s_t$ rather than thresholding isolated values.

This analysis does not claim that $A_t$ equals the final generation
error. It explains two design choices: normalizing removes trajectory
displacement as a confounding factor, and accumulating captures
unreanchored pathwise exposure. The allocation process itself is
event-triggered and therefore nonlinear: changing the local signal may
alter the first threshold-crossing position, which changes the next
cache anchor, the following reconstructed states, and all subsequent
accumulator values, so a simple local normalization can induce a much
larger downstream difference in the final trajectory.

\section{CacheScope Implementation Details}
\label{app:cachescope_impl}

The probe depth is fixed to one transformer block and
$\epsilon=10^{-6}$ in all experiments. All compared scheduling signals
share the same warm-up policy, tail policy, probe depth,
reconstruction rule, cache history, model configuration, and
generation seed.

\subsection{Threshold Calibration}
\label{app:threshold_calibration}

The threshold $\tau$ determines the average number of adaptive exact
fallbacks. Given a development set $\mathcal D_{\mathrm{dev}}$, we
select
\begin{equation}
    \tau^\star
    =
    \arg\min_{\tau}
    \left|
        \mathbb E_{\mathcal D_{\mathrm{dev}}}
        \left[
            \sum_{t=1}^{N}
            (1-b_t)\,a_t(\tau)
        \right]
        -
        B_{\mathrm{add}}
    \right|.
    \label{eq:threshold_calibration}
\end{equation}
Calibration reads only fallback counts and never any quality metric;
after calibration, $\tau^\star$ is frozen and applied to all
evaluation prompts without prompt-specific adaptation. Each backbone
and acceleration level uses an independently calibrated threshold,
since the signal scale differs across models. This procedure requires
neither training nor gradients, and avoids offline optimization of an
independent decision for every denoising step.

\subsection{Fixed Reconstructor and Cache Update}
\label{app:reconstructor_details}

At reuse steps, the skipped output is produced by the bounded
first-order residual reconstructor of the base
configuration~\cite{dicache}. Given the two most recent exact
residuals $\mathbf r_{t_1}$ and $\mathbf r_{t_2}$ with $t_1<t_2<t$,
\begin{equation}
    \widehat{\mathbf r}_t
    =
    \mathbf r_{t_2}
    +
    \gamma_t
    \left(
        \mathbf r_{t_2}
        -
        \mathbf r_{t_1}
    \right),
    \qquad
    1\leq\gamma_t\leq\gamma_{\max},
    \label{eq:app_reconstructor}
\end{equation}
with $\gamma_{\max}=1.5$. The extrapolation coefficient $\gamma_t$ is
computed from the same shallow-probe indicator following the base
reconstructor's released rule and clipped to $[1,\gamma_{\max}]$; it
is inherited unchanged and not optimized by CacheScope. If fewer than
two exact residuals are available, reconstruction falls back to
directly reusing the most recent one.

The complete per-step state transition is
\begin{equation}
\left(
    \widetilde{\mathbf r}_t,
    \mathcal H_t,
    A_t
\right)
=
\begin{cases}
\left(
    H_\theta(\mathbf y_t),
    \operatorname{Refresh}
    \left(
        \mathcal H_{t-1},
        \mathbf r_t
    \right),
    0
\right),
& e_t=1,
\\[3mm]
\left(
    \mathcal R_t(\mathcal H_{t-1}),
    \mathcal H_{t-1},
    \widetilde A_t
\right),
& e_t=0,
\end{cases}
\label{eq:complete_state_transition}
\end{equation}
where $\operatorname{Refresh}$ inserts the newly computed exact
residual into the cache history and invalidates reconstruction states
associated with the previous anchor. Only residuals from exact
evaluations enter the cache, which prevents approximate residuals from
recursively contaminating the history.

\subsection{Classifier-Free Guidance}
\label{app:cfg_details}

Under classifier-free guidance, the conditional and unconditional
branches may follow different representation trajectories. For branch
$q\in\{\mathrm{cond},\mathrm{uncond}\}$, CacheScope maintains
independent probe states, accumulators, and caches:
\begin{equation}
    s_t^{(q)}
    =
    \frac{
        \Delta y_t^{(q)}
    }{
        \Delta x_t^{(q)}+\epsilon
    },
    \qquad
    \widetilde A_t^{(q)}
    =
    A_{t-1}^{(q)}
    +
    s_t^{(q)}.
    \label{eq:branch_sensitivity}
\end{equation}
Fallback decisions and cache refreshes are applied independently per
branch, after which the outputs are combined by the original guidance
rule. This prevents the two interleaved trajectories from incorrectly
sharing probe statistics or cache anchors. On Wan2.1, the computation
budget is counted in denoiser forwards (two per sampling step), and
each branch's probe state is compared only against the previous state
of the same branch.

\subsection{Complexity}
\label{app:complexity_details}

Let $C_{\mathrm{full}}$, $C_{\mathrm{probe}}$, and $C_{\mathrm{rec}}$
denote the costs of a complete denoiser evaluation, the input
transformation plus first transformer block, and cached
reconstruction, and let $N_{\mathrm{exact}}=\sum_t e_t$. The inference
cost is approximately
\begin{equation}
\begin{aligned}
    C_{\mathrm{CacheScope}}
    \approx\;&
    N_{\mathrm{exact}}C_{\mathrm{full}}
    \\
    &+
    \left(
        N-N_{\mathrm{exact}}
    \right)
    \left(
        C_{\mathrm{probe}}
        +
        C_{\mathrm{rec}}
    \right)
    +
    \mathcal O(N).
\end{aligned}
\label{eq:computational_cost}
\end{equation}
The probe costs $3.3\%$ of a full forward on Wan2.1 and $1.8\%$ on
FLUX (Appendix~\ref{app:cost_model}). At an exact step, the probe is
part of the complete evaluation and its output is reused when
execution continues through the remaining blocks, so no network
computation is duplicated. Beyond the probe, each branch requires two
relative-change reductions, one scalar division, one scalar
accumulation, and one threshold comparison per step. The scheduler
additionally stores the previous probe input and output per branch;
the residual-cache memory is unchanged. CacheScope introduces no
trainable parameters, auxiliary predictor, gradient computation, or
offline per-timestep schedule search.

\section{Evaluation Metrics and Statistical Protocol}
\label{app:metrics_protocol}

\subsection{Baseline Configurations}
\label{app:baseline_params}

For every compared baseline, the speed-controlling hyperparameter
listed in the Param.\ column of the main-results table in the main paper
($\delta$ for TeaCache, MagCache, and DiCache, $\tau$ for EasyCache,
$N$ for TaylorSeer, and $R$ for PAB) is selected on the development
prompts so that the method's measured speedup falls inside the target
acceleration range of the corresponding regime, matching the budget of
skipped evaluations where the method's granularity allows.
Selection reads only speed and skip counts---never any quality
metric---and the chosen values are frozen before any test-set sample
is generated.
When a regime lies beyond a method's structural acceleration
ceiling---the maximum speedup its formulation can reach regardless of
the hyperparameter setting, as for MagCache on Wan2.1 Fast---the
corresponding row is marked ``--'' rather than reporting an
out-of-range configuration.
For reference, the operating points of the two released default FLUX
configurations of the DiCache baseline are reported in
Appendix~\ref{app:released_defaults}; the official defaults land at
operating points at least as aggressive as ours, so the calibration
does not weaken the baselines.

\subsection{Additional Open-Source Baseline Comparisons}
\label{app:additional_baselines}

We additionally compare CacheScope with several recently released
open-source caching methods under the same generation and evaluation
protocols. CacheScope results are reproduced for reference.

\begin{table}[t]
\centering
\scriptsize
\setlength{\tabcolsep}{4.2pt}
\renewcommand{\arraystretch}{0.88}
\caption{
\textbf{Comparison with additional open-source baselines.}
All metrics are computed against the same dense references.
CacheScope results are reproduced for comparison.
SenCache is omitted from the Wan2.1 Fast regime because its fastest
attainable configuration reaches only $2.644\times$, below the target
speedup of approximately $3.46\times$.
Best results within each backbone and regime are in bold.
}
\label{tab:additional_baselines}
\resizebox{\columnwidth}{!}{%
\begin{tabular}{lllcccc}
\toprule
Backbone & Regime & Method
& Speedup $\uparrow$
& LPIPS $\downarrow$
& PSNR $\uparrow$
& SSIM $\uparrow$ \\
\midrule

\multirow{7}{*}{Wan2.1-T2V-1.3B}
& \multirow{4}{*}{Moderate}
& SenCache
& 2.398$\times$ & 0.0775 & 25.88 & 0.8908 \\
&
& NaviCache
& 2.519$\times$ & 0.0752 & 25.72 & 0.8642 \\
&
& D2Cache
& 2.394$\times$ & 0.0933 & 23.88 & 0.8778 \\
&
& \textbf{CacheScope}
& 2.493$\times$ & \textbf{0.0414}
& \textbf{28.89} & \textbf{0.9220} \\
\cmidrule(lr){2-7}

&
\multirow{3}{*}{Fast}
& NaviCache
& 3.336$\times$ & 0.2334 & 20.81 & 0.6560 \\
&
& D2Cache
& 3.322$\times$ & 0.2075 & 19.50 & 0.7269 \\
&
& \textbf{CacheScope}
& 3.463$\times$ & \textbf{0.1607}
& \textbf{21.80} & \textbf{0.7363} \\

\midrule

\multirow{6}{*}{FLUX.1-dev}
& \multirow{3}{*}{Moderate}
& SenCache
& 2.896$\times$ & 0.1466 & 22.96 & 0.8590 \\
&
& DPCache
& 2.959$\times$ & 0.1293 & 22.58 & 0.8553 \\
&
& \textbf{CacheScope}
& 2.954$\times$ & \textbf{0.0816}
& \textbf{27.74} & \textbf{0.9219} \\
\cmidrule(lr){2-7}

&
\multirow{3}{*}{Fast}
& SenCache
& 3.494$\times$ & 0.2943 & 17.76 & 0.7482 \\
&
& DPCache
& 3.571$\times$ & 0.1562 & 21.85 & 0.8425 \\
&
& \textbf{CacheScope}
& 3.566$\times$ & \textbf{0.1246}
& \textbf{24.99} & \textbf{0.8760} \\

\midrule

\multirow{4}{*}{Open-Sora 1.2}
& \multirow{2}{*}{Moderate}
& SenCache
& 1.428$\times$ & 0.0514 & 28.44 & 0.9170 \\
&
& \textbf{CacheScope}
& 1.404$\times$ & \textbf{0.0169}
& \textbf{34.95} & \textbf{0.9623} \\
\cmidrule(lr){2-7}

&
\multirow{2}{*}{Fast}
& SenCache
& 2.138$\times$ & 0.1052 & 23.72 & 0.8586 \\
&
& \textbf{CacheScope}
& 2.052$\times$ & \textbf{0.0664}
& \textbf{26.13} & \textbf{0.9004} \\

\bottomrule
\end{tabular}%
}

\end{table}

CacheScope achieves the best fidelity metrics in every reported
backbone--regime group. On Wan2.1, it reduces LPIPS from $0.0752$ to
$0.0414$ in the Moderate regime and from $0.2075$ to $0.1607$ in the
Fast regime. On FLUX.1-dev, CacheScope also consistently improves
LPIPS, PSNR, and SSIM over SenCache and DPCache at closely matched
speedups. On Open-Sora 1.2, SenCache operates at slightly higher
speedups, but CacheScope retains substantially better fidelity in both
regimes.

\subsection{Frozen Scheduling Thresholds}

Table~\ref{tab:frozen_thresholds} lists the scheduling thresholds
calibrated on the development prompts and frozen before test-set
generation.

\begin{table}[t]
    \centering
    \small
    \caption{
    \textbf{Frozen scheduling thresholds.}
    S, D, and M differ only in their accumulated scheduling signal.
    }
    \label{tab:frozen_thresholds}
    \setlength{\tabcolsep}{5pt}
    \begin{tabularx}{\linewidth}{@{}>{\raggedright\arraybackslash}p{1.2cm}
    >{\raggedright\arraybackslash}p{1.3cm}
    >{\centering\arraybackslash}p{1.3cm}*{3}{>{\centering\arraybackslash}X}@{}}
        \toprule
        Backbone & Regime
        & Skip ratio
        & $\tau_{\mathrm{S}}$
        & $\tau_{\mathrm{D}}$
        & $\tau_{\mathrm{M}}$ \\
        \midrule
        Wan2.1
        & Moderate
        & $0.64$
        & $5.5$
        & $0.16$
        & $0.028$ \\

        Wan2.1
        & Fast
        & $0.76$
        & $20.0$
        & $0.50$
        & $0.110$ \\

        FLUX
        & Moderate
        & $0.68$
        & $4.6$
        & $0.15$
        & $0.042$ \\

        FLUX
        & Fast
        & $0.74$
        & $6.4$
        & $0.25$
        & $0.065$ \\
        \bottomrule
    \end{tabularx}
    \vspace{-4mm}
\end{table}

\subsection{Quality Metrics}

We measure fidelity relative to the corresponding full-computation
output using LPIPS~\cite{lpips}, PSNR~\cite{chen2026ntire}, and SSIM~\cite{ssim}.
LPIPS is our primary metric because it better reflects perceptual
deviation from the full-model output.

For video, LPIPS and SSIM are computed framewise and then averaged:
\begin{equation}
    \operatorname{LPIPS}_{\mathrm{video}}
    =
    \frac{1}{T}
    \sum_{i=1}^{T}
    \operatorname{LPIPS}
    \left(
        \mathbf{I}_i,
        \mathbf{I}^{\mathrm{full}}_i
    \right),
\end{equation}
and
\begin{equation}
    \operatorname{SSIM}_{\mathrm{video}}
    =
    \frac{1}{T}
    \sum_{i=1}^{T}
    \operatorname{SSIM}
    \left(
        \mathbf{I}_i,
        \mathbf{I}^{\mathrm{full}}_i
    \right).
\end{equation}

Video PSNR is computed from the global mean squared error:
\begin{equation}
    \operatorname{MSE}
    =
    \frac{1}{3THW}
    \sum_{i=1}^{T}
    \sum_{j=1}^{H}
    \sum_{k=1}^{W}
    \left\|
        \mathbf{I}_{ijk}
        -
        \mathbf{I}^{\mathrm{full}}_{ijk}
    \right\|_2^2,
\end{equation}
followed by
\begin{equation}
    \operatorname{PSNR}
    =
    10\log_{10}
    \frac{255^2}{\operatorname{MSE}}.
\end{equation}

No resizing or cropping is applied.
All compared video outputs use the same encoding and decoding pipeline.

\subsection{Efficiency Metrics}

For each sample, we record the number of exact and skipped forwards,
wall-clock latency, and actual speedup:
\begin{equation}
    \operatorname{Speedup}
    =
    \frac{
        T_{\mathrm{full}}
    }{
        T_{\mathrm{cache}}
    }.
\end{equation}

All methods are measured on identical hardware---a single NVIDIA RTX
PRO 6000D GPU---with the same model, prompt, seed, and generation
configuration.
One-time model loading is excluded.

\subsection{Statistical Protocol}

All statistical comparisons are paired at the prompt level.
For a quality metric $Q$, the paired difference between methods
$A$ and $B$ is
\begin{equation}
    d_i^{A,B}
    =
    Q_i^{A}-Q_i^{B}.
\end{equation}

LPIPS differences follow the natural convention
\begin{equation}
    d_i^{A,B}<0
    \quad\Longrightarrow\quad
    A\text{ is better}.
\end{equation}

We treat
\begin{equation}
    |d_i^{A,B}|\leq 10^{-12}
\end{equation}
as a tie and remove ties before the exact two-sided sign test.
If the resulting effective sample size satisfies
\begin{equation}
    n_{\mathrm{eff}}<8,
\end{equation}
we report the result descriptively without making a significance claim.

Confidence intervals are obtained by resampling the
prompt-level paired differences:
\begin{equation}
    \left\{
        d_i^{A,B}
    \right\}_{i=1}^{n},
\end{equation}
using $10{,}000$ bootstrap samples.
Frames are never treated as independent samples.
We apply Benjamini--Hochberg correction~\cite{bh} with
\begin{equation}
    q=0.05
\end{equation}
to the four preregistered S-versus-D LPIPS hypotheses.

Our primary analysis retains only prompt pairs for which the compared
methods execute exactly the same number of skipped forwards:
\begin{equation}
    \mathcal{I}_{A,B}
    =
    \left\{
        i
        \mid
        K_i^{A}=K_i^{B}
    \right\}.
    \label{eq:equal_skip_subset}
\end{equation}

We additionally report all $32$ prompts as a robustness check.

\section{Additional Statistical Analyses}
\label{app:additional_stats}

\subsection{Independent-Seed Replication}

To test whether the improvements depend on a particular noise
realization, we repeat the complete S-versus-D comparison with seed
$43$, without changing any threshold or hyperparameter.
Table~\ref{tab:seed_replication} reports the results.

\begin{table}[t]
    \centering
    \small
    \caption{
    \textbf{Independent-seed replication.}
    Values are median paired LPIPS differences,
    $L_{\mathrm{S}}-L_{\mathrm{D}}$, on strict matched-skip pairs.
    }
    \label{tab:seed_replication}
    \setlength{\tabcolsep}{3.5pt}
    \small
    \resizebox{\columnwidth}{!}{%
    \begin{tabular}{llccccc}
        \toprule
        Backbone
        & Regime
        & Seed 42
        & Seed 43
        & $p_{42}$
        & $p_{43}$
        & Ratio \\
        \midrule

        Wan2.1
        & Moderate
        & $-0.03353$
        & $-0.03905$
        & $9.31{\times}10^{-10}$
        & $2.16{\times}10^{-7}$
        & $1.16$ \\

        Wan2.1
        & Fast
        & $-0.05408$
        & $-0.04648$
        & $1.12{\times}10^{-7}$
        & $1.86{\times}10^{-9}$
        & $0.86$ \\

        FLUX
        & Moderate
        & $-0.05092$
        & $-0.03514$
        & $8.05{\times}10^{-7}$
        & $1.10{\times}10^{-5}$
        & $0.69$ \\

        FLUX
        & Fast
        & $-0.04578$
        & $-0.04469$
        & $1.20{\times}10^{-4}$
        & $3.05{\times}10^{-5}$
        & $0.98$ \\

        \bottomrule
    \end{tabular}}
    \vspace{-3mm}
\end{table}

The replication criteria were fixed in writing before generation:
(R1) same sign of the median paired difference, (R2) still $p<0.05$,
and (R3) median-skip range within one step, without which a cell is
treated as non-adjudicable rather than failed.
All four backbone--regime combinations satisfy R1--R3 with the
independent seed.
The seed-43/seed-42 effect-size ratios ($1.16$, $0.86$, $0.69$,
$0.98$) are reported for description only and carry no threshold;
their proximity to one suggests that the improvement is not caused by
a favorable noise initialization.

\subsection{Sensitivity to the Matching Protocol}

Strict equal-skip filtering removes prompts for which two schedulers
differ by even one exact forward.
To verify that the result is not created by this filtering, we compare
two analyses:
\begin{align}
    \mathcal{I}_{\mathrm{full}}
    &=
    \{1,\ldots,32\},
    \\
    \mathcal{I}_{\mathrm{eq}}
    &=
    \left\{
        i
        \mid
        K_i^{A}=K_i^{B}
    \right\}.
\end{align}

Across both backbones, both regimes, all three metrics, and all pairwise
scheduler comparisons, the conclusion direction does not reverse
between $\mathcal{I}_{\mathrm{full}}$ and $\mathcal{I}_{\mathrm{eq}}$.
The differences in skip count are also small:
$\operatorname{median}(K_i^{A}-K_i^{B})\in\{-1,0\}$ for all main
comparisons.
The gains therefore cannot be explained by our scheduler simply
skipping fewer denoiser evaluations.

\section{Extended Experimental Data and Audits}
\label{app:extended_data}

This appendix collects the complete measurement record underlying the
controlled scheduler comparison: signal-decorrelation statistics, the
full prompt-level paired analysis with confidence intervals and
win--tie--loss counts, the agreement of the three quality metrics, the
multiple-comparison audit, additional operating points beyond the
pre-registered levels, a schedule-placement analysis, extended
reconstruction and $\gamma_{\max}$ ablations, the analytic cost model
with its validation at $28$ operating points, a codec-independence
check, and the operating points of the released baseline defaults.
Throughout, S denotes our probe-normalized signal
$\Delta y_t/(\Delta x_t+\epsilon)$, D the default DiCache signal
$\Delta y_t$, and M the difference-based alternative
$|\Delta y_t-\Delta x_t|$; unless stated otherwise, a negative paired
median favors the first scheduler of a pair, for every metric.
All wall-clock measurements reported in this appendix were taken on a
single NVIDIA RTX PRO 6000D GPU, the same device used for the main
results.

\begin{table*}[t]
\centering
\small
\caption{\textbf{Full prompt-level paired comparison of LPIPS on the held-out
set.} ``Matched'' restricts each pair to prompts with identical skip
counts (primary read); ``all'' pools all $32$ prompts. Negative median
differences favor the first scheduler. W--T--L counts wins, ties, and
losses of the first scheduler.}
\label{tab:app_paired_full}
\setlength{\tabcolsep}{5pt}
\begin{tabularx}{\linewidth}{@{}>{\raggedright\arraybackslash}p{1.5cm}
    >{\raggedright\arraybackslash}p{1.8cm}
    >{\raggedright\arraybackslash}p{1.8cm}
    >{\raggedright\arraybackslash}p{1.5cm}
    >{\centering\arraybackslash}p{1cm}
    >{\centering\arraybackslash}p{4.3cm}*{2}{>{\centering\arraybackslash}X}@{}}
\toprule
Backbone & Level & Pair & Pairing & $n$ & Median $\Delta$ [95\% CI] & W--T--L & $p$ \\
\midrule
Wan2.1 & Moderate & S vs.\ M & matched & 30 & $-0.0051$ $[-0.0072,-0.0016]$ & 26--0--4 & $<0.001$ \\
       &          &          & all     & 32 & $-0.0062$ $[-0.0079,-0.0021]$ & 28--0--4 & $<0.001$ \\
       &          & S vs.\ D & matched & 31 & $-0.0335$ $[-0.0504,-0.0209]$ & 31--0--0 & $<0.001$ \\
       &          &          & all     & 32 & $-0.0332$ $[-0.0498,-0.0209]$ & 32--0--0 & $<0.001$ \\
       &          & M vs.\ D & matched & 29 & $-0.0273$ $[-0.0399,-0.0169]$ & 28--0--1 & $<0.001$ \\
       &          &          & all     & 32 & $-0.0273$ $[-0.0399,-0.0169]$ & 30--0--2 & $<0.001$ \\
\cmidrule(lr){2-8}
       & Fast     & S vs.\ M & matched & 32 & $-0.0634$ $[-0.0810,-0.0556]$ & 32--0--0 & $<0.001$ \\
       &          &          & all     & 32 & $-0.0634$ $[-0.0810,-0.0556]$ & 32--0--0 & $<0.001$ \\
       &          & S vs.\ D & matched & 29 & $-0.0541$ $[-0.0605,-0.0438]$ & 28--0--1 & $<0.001$ \\
       &          &          & all     & 32 & $-0.0536$ $[-0.0605,-0.0424]$ & 31--0--1 & $<0.001$ \\
       &          & M vs.\ D & matched & 29 & $+0.0098$ $[+0.0022,+0.0318]$ & 8--0--21 & $0.024$ \\
       &          &          & all     & 32 & $+0.0144$ $[+0.0022,+0.0314]$ & 9--0--23 & $0.020$ \\
\midrule
FLUX   & Moderate & S vs.\ M & matched & 23 & $-0.0126$ $[-0.0310,-0.0036]$ & 19--0--4 & $0.003$ \\
       &          &          & all     & 32 & $-0.0237$ $[-0.0326,-0.0103]$ & 28--0--4 & $<0.001$ \\
       &          & S vs.\ D & matched & 26 & $-0.0509$ $[-0.0776,-0.0357]$ & 25--0--1 & $<0.001$ \\
       &          &          & all     & 32 & $-0.0466$ $[-0.0702,-0.0315]$ & 30--0--2 & $<0.001$ \\
       &          & M vs.\ D & matched & 24 & $-0.0285$ $[-0.0466,-0.0177]$ & 20--0--4 & $0.002$ \\
       &          &          & all     & 32 & $-0.0280$ $[-0.0374,-0.0177]$ & 25--0--7 & $0.002$ \\
\cmidrule(lr){2-8}
       & Fast     & S vs.\ M & matched & 21 & $-0.0116$ $[-0.0160,-0.0070]$ & 21--0--0 & $<0.001$ \\
       &          &          & all     & 32 & $-0.0169$ $[-0.0357,-0.0120]$ & 32--0--0 & $<0.001$ \\
       &          & S vs.\ D & matched & 14 & $-0.0458$ $[-0.0714,-0.0339]$ & 14--0--0 & $<0.001$ \\
       &          &          & all     & 32 & $-0.0683$ $[-0.0803,-0.0493]$ & 32--0--0 & $<0.001$ \\
       &          & M vs.\ D & matched & 17 & $-0.0332$ $[-0.0495,-0.0254]$ & 15--0--2 & $0.002$ \\
       &          &          & all     & 32 & $-0.0369$ $[-0.0503,-0.0290]$ & 29--0--3 & $<0.001$ \\
\bottomrule
\end{tabularx}
\vspace{-4mm}
\end{table*}

\subsection{Signal Decorrelation Statistics}
\label{app:signal_stats}

The motivation for probe normalization rests on measured, not
illustrative, statistics of the two probe movements.
Over the denoising steps of the held-out runs, the Spearman rank
correlation between $\Delta y_t$ and $\Delta x_t$ is $0.95$ on
Wan2.1 and $0.52$ on FLUX: on the video backbone, almost all of the
rank variation in the raw signal is shared with input movement.
Holding $\Delta y_t$ inside its $[45,55]$ percentile band still leaves
$\Delta x_t$ spanning a factor of $1.23$ on Wan2.1 ($n{=}256$ steps in
band) and $2.45$ on FLUX ($n{=}75$), so the response ratio $s_t$ spans
a factor of $1.10$ and $2.32$ respectively (10th--90th percentile).
The consequence for scheduling is not a marginal re-ranking: ranking
the steps of a trajectory by $\Delta y_t$ and by $s_t$ and taking the
top third of each, the two sets overlap on only $4\%$ of their members
on Wan2.1 (mean over $32$ trajectories; Spearman
$r_s(\Delta y_t,s_t){=}{-}0.83$) and $37\%$ on FLUX ($16$
trajectories, $r_s{=}0.31$).
At the same skip budget, the two signals therefore spend their exact
forward passes on almost disjoint sets of steps.

\emph{Provenance.} The Wan2.1 statistics are measured on the D arm at
the Fast level, seed $42$, all $32$ held-out prompts
($\approx 2560$ steps); the FLUX statistics on the M arm at the
Moderate level, the $16$ development prompts ($\approx 688$ steps) ---
the trace dump retains $\Delta x_t$ only for those runs.
These statistics establish that the signals \emph{decide differently},
not that one decides better; $s_t$ contains $\Delta y_t$ in its
numerator, so a low overlap is expected under either hypothesis.
Which decision is better is settled by the paired analysis below.

\subsection{Full Prompt-Level Paired Analysis}
\label{app:full_paired}

Table~\ref{tab:app_paired_full} extends the main-text paired
comparison to all three scheduler pairs, under both the matched
pairing (only prompts on which the two schedulers skipped exactly the
same number of forwards; the pre-registered primary read) and the
pooled reading over all $32$ prompts (which confounds quality with
compute and is reported as a robustness check).
Confidence intervals are $95\%$ bootstrap intervals
($B{=}10^{4}$) resampling prompts, never individual frames or LPIPS
values; $p$ is a two-sided exact sign test.

Two observations beyond the main text.
First, the two comparison signals are not equivalent to each other: M
beats D in most cells but \emph{loses} at the Wan2.1 Fast level
($+0.0098$, $p{=}0.024$), where it is also the worst of the three arms
in absolute LPIPS.
Second, the matched and pooled readings never disagree in direction,
so the matched-skip restriction does not manufacture any conclusion.

\subsection{Agreement of the Three Quality Metrics}
\label{app:metric_agreement}

Table~\ref{tab:app_metric_agreement} reports every held-out cell for
all three metrics under the matched pairing.
For uniformity of reading, the sign convention is aligned across
metrics: a negative median difference means the \emph{first} scheduler
of the pair is better, for LPIPS, PSNR, and SSIM alike.
The metrics do not always agree, and we list every disagreement rather
than selecting the favourable metric per cell: in
Wan2.1/Moderate/S~vs.~M, LPIPS favors S while SSIM favors M (PSNR is a
tie); in Wan2.1/Fast/S~vs.~D, LPIPS and PSNR favor S while SSIM favors
D.
LPIPS is the pre-registered primary metric, fixed before the held-out
set was generated.

\begin{table*}[t]
\centering
\small
\caption{\textbf{Metric agreement on the held-out set, matched pairing only.}
Median paired differences; negative favors the first scheduler for all
three metrics. ``n.s.''\ denotes a sign-test tie at the $0.05$ level.}
\label{tab:app_metric_agreement}
\setlength{\tabcolsep}{4pt}
\begin{tabularx}{\linewidth}{@{}>{\raggedright\arraybackslash}p{1.5cm}
    >{\raggedright\arraybackslash}p{1.8cm}
    >{\raggedright\arraybackslash}p{1.5cm}
    >{\centering\arraybackslash}p{1cm}
  *{6}{>{\centering\arraybackslash}X}@{}}
\toprule
 & & & & \multicolumn{2}{c}{LPIPS$\downarrow$} & \multicolumn{2}{c}{PSNR$\uparrow$} & \multicolumn{2}{c}{SSIM$\uparrow$} \\
\cmidrule(lr){5-6}\cmidrule(lr){7-8}\cmidrule(lr){9-10}
Backbone & Level & Pair & $n$ & Median $\Delta$ & Verdict & Median $\Delta$ & Verdict & Median $\Delta$ & Verdict \\
\midrule
Wan2.1 & Moderate & S vs.\ M & 30 & $-0.0051$ & first & $-0.0186$ & n.s. & $+0.0122$ & second \\
       &          & S vs.\ D & 31 & $-0.0335$ & first & $-2.7257$ & first & $-0.0124$ & first \\
       &          & M vs.\ D & 29 & $-0.0273$ & first & $-3.4833$ & first & $-0.0275$ & first \\
       & Fast     & S vs.\ M & 32 & $-0.0634$ & first & $-1.6990$ & first & $-0.1107$ & first \\
       &          & S vs.\ D & 29 & $-0.0541$ & first & $-0.6389$ & first & $+0.0233$ & second \\
       &          & M vs.\ D & 29 & $+0.0098$ & second & $+1.1251$ & second & $+0.1256$ & second \\
\midrule
FLUX   & Moderate & S vs.\ M & 23 & $-0.0126$ & first & $-0.6399$ & first & $-0.0092$ & first \\
       &          & S vs.\ D & 26 & $-0.0509$ & first & $-2.9047$ & first & $-0.0494$ & first \\
       &          & M vs.\ D & 24 & $-0.0285$ & first & $-2.3443$ & first & $-0.0204$ & first \\
       & Fast     & S vs.\ M & 21 & $-0.0116$ & first & $-0.2695$ & first & $-0.0081$ & first \\
       &          & S vs.\ D & 14 & $-0.0458$ & first & $-3.9094$ & first & $-0.0381$ & first \\
       &          & M vs.\ D & 17 & $-0.0332$ & first & $-2.5953$ & first & $-0.0324$ & first \\
\bottomrule
\end{tabularx}
\vspace{-4mm}
\end{table*}

Counting every held-out cell in which S participates across the two
backbones, two levels, three metrics, and two pairs, there are $24$
cells: S is significantly better in $21$, significantly worse in $2$
(both SSIM, on Wan2.1), and indistinguishable in $1$ (PSNR,
Wan2.1/Moderate/S~vs.~M).

\subsection{VBench Sub-Dimension Evaluation on Wan2.1}
\label{app:vbench_eval}

The fidelity metrics of the main paper measure agreement with the
full-computation output. As a complementary reference-free view,
Table~\ref{tab:app_vbench} reports raw VBench sub-dimension scores on
a subset of VBench prompts with Wan2.1 as the backbone, comparing the
caching methods at the Moderate operating point against the
full-computation (dense) baseline.
CacheScope obtains the best score among the caching methods on all
seven quality dimensions, and matches the dense baseline's dynamic
degree exactly; Panel~(b) of the reconstruction-ablation figure in the main paper visualizes the normalized profiles.

\begin{table}[t]
\centering
\small
\caption{\textbf{Raw VBench sub-dimension scores on Wan2.1
(Moderate regime).} Higher is better for all dimensions except
dynamic degree (Dyn.), where the dense baseline's value is the
reference and deviations in either direction indicate drift; Dyn.\ is
therefore excluded from the bolding. Best value per column among the
caching methods is in bold. Subj.\ $=$ subject consistency, BG $=$
background consistency, Flick.\ $=$ temporal flickering, Smooth $=$
motion smoothness, Aes.\ $=$ aesthetic quality, Imag.\ $=$ imaging
quality, Overall $=$ overall consistency.}
\label{tab:app_vbench}
\setlength{\tabcolsep}{2pt}
\resizebox{\columnwidth}{!}{
\begin{tabular}{lcccccccc}
\toprule
Method & Subj. & BG & Flick. & Smooth & Aes. & Imag. & Overall & Dyn. \\
\midrule
Wan2.1 (dense) & 0.9723 & 0.9791 & 0.9843 & 0.9905 & 0.6152 & 0.7155 & 0.1974 & 0.2500 \\
\midrule
TeaCache & 0.9710 & 0.9783 & 0.9847 & 0.9898 & 0.6225 & 0.7051 & 0.1976 & 0.2812 \\
MagCache & 0.9721 & 0.9790 & 0.9843 & 0.9898 & 0.6081 & 0.7091 & 0.1981 & 0.2188 \\
EasyCache & 0.9705 & 0.9761 & 0.9841 & 0.9901 & 0.5903 & 0.7127 & 0.1980 & 0.1875 \\
DiCache & 0.9719 & 0.9771 & 0.9843 & 0.9901 & 0.6128 & 0.7117 & 0.1956 & 0.2500 \\
\textbf{CacheScope} & \textbf{0.9813} & \textbf{0.9970} & \textbf{0.9852} & \textbf{0.9902} & \textbf{0.6274} & \textbf{0.7133} & \textbf{0.1986} & 0.2500 \\
\bottomrule
\end{tabular}
}
\vspace{-4mm}
\end{table}

\subsection{Multiple-Comparison Audit}
\label{app:multiplicity}

Two Benjamini--Hochberg families at $q{=}0.05$ were declared before
the results were examined: a \emph{confirmatory} family containing
only the four pre-registered hypotheses (S~vs.~D, LPIPS, matched
pairing; one cell per backbone and level), and an \emph{exploratory}
family containing every other reported matched cell
($m{=}77$).
The confirmatory family survives the correction entirely.
In the exploratory family, $62$ of $77$ cells are nominally
significant at the raw $0.05$ level and $61$ survive the correction.

Table~\ref{tab:app_bh_unrejected} lists all $16$ unrejected
exploratory cells.
Fifteen of them were never significant to begin with
($0.071\le p\le 1.000$); exactly one cell is nominally significant and
then lost to the correction, namely
Wan2.1/FLUX-matched/S~vs.~D/SSIM
($n{=}30$, median $-0.0318$, $p{=}0.043$).
Nine further cells have fewer than eight identical-skip prompts and
are excluded from the family (reported but not tested):
Wan2.1/Aggressive/M~vs.~D (LPIPS, PSNR, SSIM; $n{=}4$ each),
Wan2.1/Conservative/S~vs.~M (LPIPS, PSNR, SSIM; $n{=}5$ each), and
Wan2.1/Conservative/M~vs.~D (LPIPS, PSNR, SSIM; $n{=}1$ each).

\begin{table}[t]
\centering
\small
\caption{\textbf{The $16$ exploratory cells not rejected under
Benjamini--Hochberg ($q{=}0.05$).} Only the first was nominally
significant before correction.}
\label{tab:app_bh_unrejected}
\setlength{\tabcolsep}{2.5pt}
\resizebox{\columnwidth}{!}{
\begin{tabular}{llcrr}
\toprule
Cell & Metric & $n$ & Median & $p$ \\
\midrule
Wan2.1, FLUX-matched, S vs.\ D & SSIM & 30 & $-0.03180$ & $0.043$ \\
FLUX, ret0.2-Moderate, M vs.\ D & PSNR & 31 & $-0.06527$ & $0.071$ \\
Wan2.1, FLUX-matched, S vs.\ M & LPIPS & 21 & $-0.01347$ & $0.078$ \\
Wan2.1, FLUX-matched-lo, S vs.\ M & PSNR & 28 & $-0.11398$ & $0.087$ \\
Wan2.1, FLUX-matched-lo, S vs.\ M & SSIM & 28 & $+0.00621$ & $0.087$ \\
FLUX, Aggressive, S vs.\ M & PSNR & 19 & $-0.16893$ & $0.167$ \\
Wan2.1, FLUX-matched, S vs.\ M & PSNR & 21 & $-0.50966$ & $0.189$ \\
FLUX, Aggressive, S vs.\ M & SSIM & 19 & $+0.00866$ & $0.359$ \\
FLUX, Aggressive, S vs.\ D & LPIPS & 19 & $-0.03466$ & $0.359$ \\
Wan2.1, Aggressive, S vs.\ D & SSIM & 21 & $-0.02356$ & $0.383$ \\
Wan2.1, FLUX-matched, S vs.\ M & SSIM & 21 & $+0.00466$ & $0.383$ \\
FLUX, ret0.2-Moderate, M vs.\ D & LPIPS & 31 & $+0.00065$ & $0.720$ \\
FLUX, ret0.2-Moderate, M vs.\ D & SSIM & 31 & $+0.00036$ & $0.720$ \\
Wan2.1, Moderate, S vs.\ M & PSNR & 30 & $-0.01858$ & $1.000$ \\
FLUX, Aggressive, S vs.\ M & LPIPS & 19 & $-0.00144$ & $1.000$ \\
FLUX, Aggressive, S vs.\ D & SSIM & 19 & $-0.00290$ & $1.000$ \\
\bottomrule
\end{tabular}
}
\vspace{-3.5mm}
\end{table}

Two further audits.
\emph{Family-boundary sensitivity:} restricting the exploratory family
to the two pre-registered levels ($m{=}32$) rejects $31$ cells;
including every post-hoc level ($m{=}77$, as done here) rejects $61$;
no cell rejected under the narrower family fails under the wider one.
\emph{Margins:} the tightest surviving cell
(Wan2.1/Aggressive/S~vs.~M/SSIM) has $p{=}0.039$ against a BH
criterion of $0.0396$ --- a headroom factor of only $1.01$, which we
state rather than describing the family as comfortably surviving.
Among rejected cells, the one most affected by the video codec
(Section~\ref{app:codec}) is Wan2.1/Fast/M~vs.~D/LPIPS: its paired
median is $+0.00976$ under the H.264 pipeline but $+0.02048$ on raw
tensors, a residual drift of $52\%$ of the effect; it clears its BH
criterion by a factor of $1.6$ and the adjudication is the same either
way, but this is the one cell that is thin on both margins at once.

\subsection{Dependence on the Skip Ratio}
\label{app:skip_ratio}

Table~\ref{tab:app_skip_ladder} traces the paired LPIPS gaps as the
skip ratio $\rho$ grows, including post-hoc levels beyond the
pre-registered pair (marked $\dagger$; all folded into the exploratory
BH family).
Within a backbone, rows sharing a retention ratio are directly
comparable and may be read as a curve; levels with a different
retention ratio are listed separately because their warm-up lengths
differ.

\begin{table}[t]
\centering
\small
\caption{\textbf{Paired LPIPS medians on identical-skip prompts as a function
of the skip ratio $\rho$; negative favors the first scheduler, $n$ in
parentheses.} Cells with $n<8$ are below the pre-registered minimum and
reported without adjudication. $\dagger$ marks post-hoc levels.}
\label{tab:app_skip_ladder}
\setlength{\tabcolsep}{2pt}
\resizebox{\columnwidth}{!}{
\begin{tabular}{llcccc}
\toprule
Backbone & Level & $\rho$ & S vs.\ D & S vs.\ M & M vs.\ D \\
\midrule
\multicolumn{6}{@{}l}{\emph{Retention $0.2$ (Wan2.1 main-table setting):}} \\
Wan2.1 & Conservative$^\dagger$ & 0.52 & $-0.0154$ (28) & $n{=}5$ & $n{=}1$ \\
Wan2.1 & Moderate & 0.64 & $-0.0335$ (31) & $-0.0051$ (30) & $-0.0273$ (29) \\
Wan2.1 & Fast & 0.76 & $-0.0541$ (29) & $-0.0634$ (32) & $+0.0098$ (29) \\
\midrule
\multicolumn{6}{@{}l}{\emph{Retention $0.1$ (matched to FLUX):}} \\
Wan2.1 & FLUX-matched-lo$^\dagger$ & 0.68 & $-0.0729$ (31) & $-0.0058$ (28) & $-0.0612$ (27) \\
Wan2.1 & FLUX-matched$^\dagger$ & 0.74 & $-0.0800$ (30) & $-0.0135$ (21) & $-0.0624$ (20) \\
Wan2.1 & Aggressive$^\dagger$ & 0.82 & $-0.0809$ (21) & $-0.0619$ (9) & $n{=}4$ \\
\midrule
\multicolumn{6}{@{}l}{\emph{Retention $0.1$ (FLUX main-table setting):}} \\
FLUX & Moderate & 0.68 & $-0.0509$ (26) & $-0.0126$ (23) & $-0.0285$ (24) \\
FLUX & Fast & 0.74 & $-0.0458$ (14) & $-0.0116$ (21) & $-0.0332$ (17) \\
FLUX & Aggressive$^\dagger$ & 0.82 & $-0.0347$ (19) & $-0.0014$ (19) & $-0.0566$ (32) \\
\midrule
\multicolumn{6}{@{}l}{\emph{Retention $0.2$ on FLUX (warm-up control):}} \\
FLUX & ret0.2-Moderate$^\dagger$ & 0.68 & $-0.0155$ (32) & $-0.0146$ (31) & $+0.0006$ (31) \\
\bottomrule
\end{tabular}
}
\vspace{-3.5mm}
\end{table}

The two backbones behave oppositely as $\rho$ grows.
Against D, the advantage of S widens monotonically on Wan2.1
($-0.0154\to-0.0335\to-0.0541$ across $\rho=0.52,0.64,0.76$) and
shrinks monotonically on FLUX
($-0.0509\to-0.0458\to-0.0347$ across $\rho=0.68,0.74,0.82$, the last
no longer significant); against M, the FLUX gap closes entirely at
$\rho{=}0.82$ ($-0.0014$, $n{=}19$, $p{=}1.000$).
Because the retention ratio moves together with the backbone in the
pre-registered levels ($0.2$ on Wan2.1, $0.1$ on FLUX), we ran the
levels that separate the two factors.
With retention held at $0.1$ and the same three skip ratios on both
backbones, the opposite trends persist
(Wan2.1: $-0.0729\to-0.0800\to-0.0809$, all significant; FLUX:
$-0.0509\to-0.0458\to-0.0347$, significant at $\rho{=}0.68,0.74$
only), and at fixed $\rho{=}0.68$ on FLUX, varying only the retention
ratio moves the S-vs.-D gap from $-0.0509$ (retention $0.1$) to
$-0.0155$ (retention $0.2$), both significant.
We therefore state the conclusion as \emph{probe-signal construction
is backbone-dependent} and do not claim a universal ordering at high
skip ratios.

\subsection{Additional Operating Points}
\label{app:additional_levels}

Table~\ref{tab:app_extra_levels} reports the six post-hoc levels in
full.
Each was chosen for a stated, exogenous reason: Conservative extends
the pre-registered Wan2.1 ladder one spacing ($12$ skipped forwards)
downward; FLUX/Aggressive coincides with the released default
operating point of the official M configuration
(Section~\ref{app:released_defaults}); Wan2.1/Aggressive matches the
$\rho$ at which the S-vs.-M gap closes on FLUX; FLUX-matched and
FLUX-matched-lo place Wan2.1 on FLUX's retention and skip ratios; and
ret0.2-Moderate holds the FLUX backbone and skip ratio fixed while
raising only the retention ratio.
Thresholds for every level were read off the development set and
frozen, together with the decision rule, before generation.

\begin{table*}[t]
\centering
\small
\caption{\textbf{Post-hoc operating points (held-out set, $n{=}32$ prompts,
seed $42$).} $\bar{s}$ is the median skip count. The matched-compute
gate passes when the median-skip spread across the three schedulers is
zero. Paired LPIPS rows use the matched pairing; cells with $n<8$ are
reported without adjudication. For Wan2.1/Aggressive, the M arm was
calibrated to $82$ on the development set but reached $83.5$ on the
held-out set (calibration drift); for Wan2.1/Conservative, no
threshold places M at skip $52$ because its median skip jumps from
$49$ to $54$ between adjacent threshold values on the development
grid.}
\label{tab:app_extra_levels}
\setlength{\tabcolsep}{3.6pt}
\begin{tabularx}{\linewidth}{@{}>{\raggedright\arraybackslash}p{3.5cm}
    >{\centering\arraybackslash}p{1cm}
    >{\centering\arraybackslash}p{1cm}*{4}{>{\centering\arraybackslash}X} 
    >{\raggedright\arraybackslash}p{6.5cm}@{}}
\toprule
Level & Scheduler & $\bar{s}$ & LPIPS$\downarrow$ & PSNR$\uparrow$ & Speedup & Gate & Paired LPIPS (matched) \\
\midrule
Wan2.1 Conservative$^\dagger$ & S & 52 & 0.0169 & 32.24 & 1.95$\times$ & \multirow{3}{*}{FAIL} &
S vs.\ D: $n{=}28$, $-0.01539$, 23--0--5, $p{<}0.001$ \\
($\rho{=}0.52$, ret.\ $0.2$) & D & 52 & 0.0429 & 29.10 & 1.95$\times$ & &
S vs.\ M: $n{=}5$, $-0.00795$, 5--0--0, $p{=}0.062$ (n.a.) \\
 & M & 54 & 0.0284 & 30.56 & 2.02$\times$ & &
M vs.\ D: $n{=}1$, $-0.03604$, 1--0--0, $p{=}1.000$ (n.a.) \\
\midrule
Wan2.1 FLUX-matched-lo$^\dagger$ & S & 68 & 0.1041 & 24.05 & 2.66$\times$ & \multirow{3}{*}{PASS} &
S vs.\ D: $n{=}31$, $-0.07290$, 28--0--3, $p{<}0.001$ \\
($\rho{=}0.68$, ret.\ $0.1$) & D & 68 & 0.1787 & 21.08 & 2.75$\times$ & &
S vs.\ M: $n{=}28$, $-0.00576$, 21--0--7, $p{=}0.013$ \\
 & M & 68 & 0.0963 & 23.81 & 2.75$\times$ & &
M vs.\ D: $n{=}27$, $-0.06117$, 27--0--0, $p{<}0.001$ \\
\midrule
Wan2.1 FLUX-matched$^\dagger$ & S & 74 & 0.1310 & 22.63 & 3.25$\times$ & \multirow{3}{*}{PASS} &
S vs.\ D: $n{=}30$, $-0.08004$, 25--0--5, $p{<}0.001$ \\
($\rho{=}0.74$, ret.\ $0.1$) & D & 74 & 0.2124 & 19.66 & 3.25$\times$ & &
S vs.\ M: $n{=}21$, $-0.01347$, 15--0--6, $p{=}0.078$ \\
 & M & 74 & 0.1484 & 22.40 & 3.25$\times$ & &
M vs.\ D: $n{=}20$, $-0.06237$, 19--0--1, $p{<}0.001$ \\
\midrule
Wan2.1 Aggressive$^\dagger$ & S & 82 & 0.2209 & 20.24 & 4.30$\times$ & \multirow{3}{*}{FAIL} &
S vs.\ D: $n{=}21$, $-0.08089$, 21--0--0, $p{<}0.001$ \\
($\rho{=}0.82$, ret.\ $0.1$) & D & 82 & 0.3071 & 18.38 & 4.30$\times$ & &
S vs.\ M: $n{=}9$, $-0.06186$, 9--0--0, $p{=}0.004$ \\
 & M & 83.5 & 0.2940 & 18.13 & 4.57$\times$ & &
M vs.\ D: $n{=}4$, $+0.15750$, 2--0--2, $p{=}1.000$ (n.a.) \\
\midrule
FLUX Aggressive$^\dagger$ & S & 41 & 0.3394 & 19.98 & 4.93$\times$ & \multirow{3}{*}{PASS} &
S vs.\ D: $n{=}19$, $-0.03466$, 12--0--7, $p{=}0.359$ \\
($\rho{=}0.82$, ret.\ $0.1$) & D & 41 & 0.3620 & 18.96 & 4.94$\times$ & &
S vs.\ M: $n{=}19$, $-0.00144$, 10--0--9, $p{=}1.000$ \\
 & M & 41 & 0.3315 & 19.96 & 4.95$\times$ & &
M vs.\ D: $n{=}32$, $-0.05662$, 30--0--2, $p{<}0.001$ \\
\midrule
FLUX ret0.2-Moderate$^\dagger$ & S & 34 & 0.1084 & 27.34 & 2.96$\times$ & \multirow{3}{*}{PASS} &
S vs.\ D: $n{=}32$, $-0.01545$, 26--0--6, $p{<}0.001$ \\
($\rho{=}0.68$, ret.\ $0.2$) & D & 34 & 0.1285 & 25.40 & 2.95$\times$ & &
S vs.\ M: $n{=}31$, $-0.01462$, 27--0--4, $p{<}0.001$ \\
 & M & 34 & 0.1337 & 25.57 & 2.95$\times$ & &
M vs.\ D: $n{=}31$, $+0.00065$, 14--0--17, $p{=}0.720$ \\
\bottomrule
\end{tabularx}
\vspace{-4mm}
\end{table*}

For the retention-$0.1$ Wan2.1 levels and ret0.2-Moderate, the
threshold-selection rule differs from the other levels: instead of
minimizing the gap between the median skip and the target, it
maximizes the number of prompts whose skip equals the target exactly
(ties broken by median gap, then by the smaller threshold).
This rule was adopted after an initial calibration returned arms whose
median skips agreed to within half a step yet whose primary pair
coincided on only $7$ of $16$ development prompts; the revised rule
was fixed before the re-scan, depends only on skip counts, and no
quality metric was read at any point during calibration.

\subsection{Schedule Placement Analysis}
\label{app:placement}

By construction the three schedulers skip the same number of steps, so
any quality difference must come from the \emph{placement} of the
exact forwards.
Table~\ref{tab:app_placement} measures that placement on FLUX at a
matched skip budget ($\rho{=}0.68$) under two retention settings ---
retention $0.1$ leaves $8$--$10$ forward passes free to place and
retention $0.2$ leaves $4$, so this is a minimal manipulation of
warm-up length alone.
The centroid is the mean position of the kept passes within the
keepable window, normalized to $[0,1]$; $\Delta_{\mathrm{cen}}$ is the
absolute centroid gap between two schedulers; $d_{\mathrm{NN}}$ is
their symmetric nearest-neighbour distance in forward passes; and
$\mathbb{E}[d]$ is its exact expectation under a null model that keeps
one arm's actual choice and redraws the other uniformly, so
$d_{\mathrm{NN}}/\mathbb{E}[d]<1$ means the two schedules agree more
than chance.

\begin{table}[t]
\centering
\small
\caption{\textbf{Where each scheduler spends its exact forward passes (FLUX,
$\rho{=}0.68$).} Top: centroids of the kept passes. Bottom: pairwise
placement statistics; $\Delta$LPIPS repeats the held-out paired median
for alignment. Mechanism columns are measured on the $16$ development
prompts, the only runs with per-step traces.}
\label{tab:app_placement}
\setlength{\tabcolsep}{3.5pt}
\begin{tabularx}{\linewidth}{@{}>{\raggedright\arraybackslash}p{3.5cm}*{3}{>{\centering\arraybackslash}X} 
    >{\raggedright\arraybackslash}p{6.5cm}@{}}
\toprule
Retention & S & D & M \\
\midrule
$r=0.1$ (8--10 free) & 0.423 & 0.570 & 0.471 \\
$r=0.2$ (4 free)     & 0.449 & 0.635 & 0.595 \\
\bottomrule
\end{tabularx}

\vspace{2mm}

\resizebox{\columnwidth}{!}{
\begin{tabularx}{\linewidth}{@{}>
{\raggedright\arraybackslash}p{0.5cm}>{\raggedright\arraybackslash}p{1.2cm}*{2}{>{\centering\arraybackslash}X} 
>{\centering\arraybackslash}p{1.2cm}>{\centering\arraybackslash}p{1.9cm}
    >{\centering\arraybackslash}p{0.5cm}@{}}
\toprule
$r$ & Pair & $\Delta_{\mathrm{cen}}$ & $d_{\mathrm{NN}}$ &
$d_{\mathrm{NN}}/\mathbb{E}[d]$ & $\Delta$LPIPS & $n$ \\
\midrule
0.1 & S vs.\ D & 0.147 & 1.50 & 0.723 & $-0.0509$ & 26 \\
    & S vs.\ M & 0.048 & 1.19 & 0.560 & $-0.0126$ & 23 \\
    & M vs.\ D & 0.099 & 1.22 & 0.574 & $-0.0285$ & 24 \\
0.2 & S vs.\ D & 0.186 & 2.25 & 0.553 & $-0.0155$ & 32 \\
    & S vs.\ M & 0.145 & 2.75 & 0.668 & $-0.0146$ & 31 \\
    & M vs.\ D & 0.041 & 2.25 & 0.538 & $+0.0006$ (n.s.) & 31 \\
\bottomrule
\end{tabularx}
}
\vspace{-5mm}
\end{table}

Three readings.
First, S spends its budget noticeably earlier in the keepable window
(centroid $0.423$ versus $0.570$ for D at retention $0.1$).
Second, all six $d_{\mathrm{NN}}/\mathbb{E}[d]$ ratios are below one
($0.54$--$0.72$): the three signals concur on roughly where the
sensitive region is and differ in phase within it, so the disagreement
is systematic rather than noise.
Third, raising the retention squeezes M onto D --- their centroid gap
falls from $0.099$ to $0.041$, exactly where their paired LPIPS
difference loses significance ($-0.0285\to+0.0006$) --- while S stays
apart from D ($0.147\to0.186$) and retains a significant advantage.
A signal that is a difference of two movements degenerates towards its
dominant term when the schedule has few degrees of freedom; a signal
that is a ratio does not.

\subsection{Offline Signal Aggregation and Schedule Information}
\label{app:offline_schedule_information}

\paragraph{Motivation and construction.}
We further examine whether the advantage of the trajectory-normalized
signal relies on prompt-specific online variation, or whether it also
reveals a stable timestep structure shared across prompts.
We construct two fixed schedules using only the $16$-prompt development
set.
For \textbf{Global-S}, we average
$s_t=\Delta y_t/(\Delta x_t+\epsilon)$ independently at each timestep
and apply the same accumulator-and-threshold rule used by CacheScope to
the resulting mean curve.
For \textbf{Global-D}, we repeat exactly the same procedure using the
raw probe variation $\Delta y_t$.
The thresholds
$\tau_{\mathrm{S}}=4.5267$ and $\tau_{\mathrm{D}}=0.1405$
are selected solely to obtain the target skip count of $34$;
no fidelity metric is used during schedule construction.

Both fixed masks are generated before any held-out output or metric is
accessed, and the development and held-out prompt sets are disjoint.
We additionally evaluate \textbf{cross-prompt replay}, in which the
Online-S schedule generated for prompt $i$ is applied to prompt $i+1$
under a cyclic derangement with no fixed points.
All four variants continue to execute the same shallow probe; only the
exact/skip decisions are replaced.
The comparison therefore isolates the information supplied by the
scheduling signal and by prompt-specific schedule assignment.

\begin{table*}[t]
\centering
\small
\caption{\textbf{Offline signal aggregation and schedule-information
analysis on FLUX.1-dev at the Moderate operating point
($\rho{=}0.68$).}
Global-S and Global-D are constructed from the same $16$-prompt
development set using the same timestep aggregation, threshold solver,
and skip budget; they differ only in the aggregated scheduling signal.
Quality values are medians over $32$ held-out prompts.
The paired LPIPS difference is
$\Delta L=L_{\mathrm{variant}}-L_{\mathrm{Online\text{-}S}}$;
positive values favor Online-S.
The final column reports an exact two-sided sign test against Online-S.}
\label{tab:app_offline_schedule_information}
\setlength{\tabcolsep}{5pt}
\resizebox{\textwidth}{!}{
\begin{tabular}{llcccccc}
\toprule
Variant
& Signal or schedule source
& Prompt-specific
& Exact fwd.\ total
& LPIPS $\downarrow$
& PSNR $\uparrow$
& SSIM $\uparrow$
& Median $\Delta L$; $p$ \\
\midrule
Online-S
& $s_t$, evaluated online
& Yes
& 515
& 0.0816
& 27.74
& 0.9219
& Reference \\

Global-S
& Mean $s_t$ on development set
& No
& 512
& 0.0792
& 27.54
& 0.9202
& $+0.0000$; $p{=}1.000$ \\

Global-D
& Mean $\Delta y_t$ on development set
& No
& 512
& 0.1337
& 23.31
& 0.8734
& $+0.0437$; $p{=}2.6{\times}10^{-6}$ \\

Cross-prompt replay
& Online-S mask from another prompt
& Mismatched
& 515
& 0.0794
& 27.35
& 0.9201
& $+0.0000$; $p{=}1.000$ \\
\bottomrule
\end{tabular}
}
\vspace{-2mm}
\end{table*}

\paragraph{Results.}
Global-S is statistically indistinguishable from Online-S:
its median LPIPS is $0.0792$, compared with $0.0816$ for Online-S,
with $14$ improvements, $13$ degradations, and $5$ ties across the
$32$ paired prompts ($p{=}1.000$).
Cross-prompt replay is likewise statistically indistinguishable from
Online-S.
Thus, at this FLUX operating point, a substantial component of the
useful scheduling structure is shared across prompts, while
prompt-specific schedule variation provides no additional
statistically detectable benefit~\cite{zhang2025matagent,zhang2025failure}.
This result is specific to the evaluated backbone and operating point
and does not imply that online adaptation is unnecessary in general~\cite{zhang2026agora}.

Importantly, applying the same offline construction to raw probe
variation does not recover the same quality.
Global-D raises median LPIPS to $0.1337$ and is worse than Online-S on
$29$ of $32$ prompts, with a median paired increase of $0.0437$
($p{=}2.6{\times}10^{-6}$).
Global-S and Global-D use identical development prompts, aggregation
procedures, threshold solvers, and skip budgets; the only changed
quantity is whether the probe response is normalized by the
corresponding input movement.
This controlled comparison therefore supports the central claim that
trajectory normalization exposes a substantially more effective
timestep-importance profile than raw probe-output variation.

The fixed schedules execute $512$ exact forwards in total, compared
with $515$ for Online-S, because Online-S produces per-prompt skip
counts of $33$, $34$, or $35$, whereas a shared mask must use a single
count.
The $0.6\%$ difference is conservative for the fixed schedules and
cannot explain their comparable fidelity.

\paragraph{Schedule diversity.}
The online schedules are not identical: the $32$ Online-S runs produce
nine distinct masks, with a median pairwise Jaccard similarity of
$0.200$ within the adaptive window.
Nevertheless, cross-prompt replay does not cause a detectable quality
loss at this operating point, suggesting that some of this mask
variation is functionally interchangeable.
Global-S and Global-D have a Jaccard similarity of only $0.286$.
Their exact-step sets within the adaptive window are
\[
\{7,9,12,16,19,25,32,38,44\}
\]
for Global-S and
\[
\{9,14,19,26,32,37,41,44,47\}
\]
for Global-D.
The large fidelity difference therefore accompanies a substantial
change in exact-compute placement rather than a nearly identical mask.

\subsection{Extended Reconstruction Ablation}
\label{app:reconstruction_extended}

Table~\ref{tab:app_reconstruction_full}
complements the reconstruction-ablation figure in the main paper
with absolute median LPIPS on both
backbones and win--loss counts
against the common reference (least squares, $k{=}2$).
All arms replay an identical exact/skip decision sequence, verified by
comparing an MD5 hash of that sequence prompt by prompt ($32/32$
agreeing on FLUX, $32/32$ on Wan2.1), so the arms are exactly
compute-matched and every difference comes from the reconstruction
formula alone.

\begin{table}[t]
\centering
\small
\caption{\textbf{Median LPIPS of five cached-residual reconstructors on the
$16$-prompt development set, with W--L counts against the least
squares $k{=}2$ reference.} All arms replay identical exact/skip
schedules.}
\label{tab:app_reconstruction_full}
\setlength{\tabcolsep}{2pt}
\resizebox{\columnwidth}{!}{
\begin{tabular}{lcccccccc}
\toprule
 & \multicolumn{4}{c}{Wan2.1} & \multicolumn{4}{c}{FLUX} \\
\cmidrule(lr){2-5}\cmidrule(lr){6-9}
Reconstructor & Mod. & W--L & Fast & W--L & Mod. & W--L & Fast & W--L \\
\midrule
Direct reuse & 0.0823 & 16--0 & 0.1314 & 15--1 & 0.1634 & 0--16 & 0.4020 & 8--8 \\
$\gamma$ extrap.\ (used) & \textbf{0.0357} & 16--0 & \textbf{0.0747} & 16--0 & \textbf{0.1170} & 16--0 & \textbf{0.3379} & 14--2 \\
LS, $k{=}1$ & 0.0812 & -- & 0.1312 & -- & -- & -- & -- & -- \\
LS, $k{=}2$ & 0.3128 & (ref) & 0.5273 & (ref) & 0.1295 & (ref) & 0.3770 & (ref) \\
LS, $k{=}4$ & 0.6670 & 1--15 & 0.5894 & 6--10 & 0.1266 & 7--9 & 0.4537 & 1--15 \\
LS, $k{=}8$ & 0.7036 & 1--15 & 0.7280 & 2--14 & 0.1311 & 3--13 & 0.4839 & 0--16 \\
\bottomrule
\end{tabular}
}\vspace{-4mm}
\end{table}

Two things are visible.
First, the bounded first-order $\gamma$ extrapolation is at least as
good as every alternative in all four columns.
Second, extending the history is monotonically worse on both
backbones, and on Wan2.1 it is catastrophic
($0.3128\to0.6670\to0.7036$ at the Moderate level): more history means
fitting a smooth model to a residual sequence that is not smooth over
that horizon, and on video the extrapolation diverges.
We report this as a bounded negative result, not a theorem: it rests
on $16$ development prompts and a single seed, and at $n{=}16$ a
two-sided exact sign test needs $13$--$3$ to reach $p{=}0.021$, so the
table establishes an ordering that is stable across two backbones and
two levels rather than adjudicating significance.

\subsection{Reconstruction Cap Ablation}
\label{app:gamma_cap}

Table~\ref{tab:app_gamma_cap} sweeps the reconstruction gain cap
$\gamma_{\max}$ on the development set (sweeping a hyperparameter on
the held-out set would forfeit its untouched status).
Differences are paired against $\gamma_{\max}{=}1.5$ and restricted to
prompts with identical skip counts; negative is better.

\begin{table}[t]
\centering
\small
\caption{\textbf{Ablation of $\gamma_{\max}$ on the $16$-prompt development
set, paired against the default $\gamma_{\max}{=}1.5$.} No cell reaches
significance (smallest of the $12$ $p$-values: $0.077$, before any
correction).}
\label{tab:app_gamma_cap}
\setlength{\tabcolsep}{4pt}

\begin{tabularx}{\linewidth}{@{}>{\raggedright\arraybackslash}p{1.2cm}
>{\raggedright\arraybackslash}p{1.4cm}*{2}{>{\centering\arraybackslash}X} 
>{\centering\arraybackslash}p{2.2cm}
    >{\centering\arraybackslash}p{0.9cm}@{}}
\toprule
Scheduler & Level & $\gamma_{\max}$ & $n_{\mathrm{eff}}$ &
Median $\Delta$LPIPS & $p$ \\
\midrule
S & Fast & 2.0 & 16 & $-0.00453$ & 0.077 \\
S & Fast & 3.0 & 15 & $+0.00392$ & 1.000 \\
S & Fast & $\infty$ & 15 & $+0.00519$ & 1.000 \\
S & Aggressive & 2.0 & 16 & $+0.00612$ & 0.804 \\
S & Aggressive & 3.0 & 14 & $+0.01785$ & 0.180 \\
S & Aggressive & $\infty$ & 13 & $+0.02660$ & 0.267 \\
M & Fast & 2.0 & 14 & $+0.00020$ & 1.000 \\
M & Fast & 3.0 & 15 & $+0.01326$ & 0.118 \\
M & Fast & $\infty$ & 15 & $+0.01576$ & 0.118 \\
M & Aggressive & 2.0 & 16 & $-0.00103$ & 1.000 \\
M & Aggressive & 3.0 & 16 & $+0.00798$ & 0.455 \\
M & Aggressive & $\infty$ & 16 & $+0.03819$ & 0.077 \\
\bottomrule
\end{tabularx}
\vspace{-1mm}
\end{table}

\begin{table}[t]
\centering
\small
\caption{\textbf{Per-forward cost from the architecture specification.}
``Non-skip.''\ counts forwards the schedule can never skip (warm-up
$+$ forced tail).}
\label{tab:app_cost_model}
\setlength{\tabcolsep}{2pt}
\resizebox{\columnwidth}{!}{
\begin{tabular}{lcccccc}
\toprule
Backbone & Blocks & Tok./fwd & $N$ & Non-skip. &
Exact fwd.\ FLOPs & Probe fwd.\ FLOPs \\
\midrule
Wan2.1 & 30 & 32760 & 100 & 20 & $2.83{\times}10^{14}$ &
$9.43{\times}10^{12}$ (3.3\%) \\
FLUX & 19+38 & 4608 & 50 & 6+1 & $7.44{\times}10^{13}$ &
$1.30{\times}10^{12}$ (1.8\%) \\
\bottomrule
\end{tabular}
}
\vspace{-3mm}
\end{table}

The two cells tied at the smallest $p$-value point in opposite
directions (one toward a looser cap, one toward the default), which is
what an inert hyperparameter looks like.
We therefore read the cap as inert over the swept range and keep
$\gamma_{\max}{=}1.5$ because it is the released default, not because
it won.

\begin{table}[t]
\centering
\small
\caption{\textbf{Analytic versus measured speedup at every skip count with at
least eight prompts, pooled across all three schedulers (the cost
model depends only on the skip count).}
$S_{\mathrm{pred}}$ is re-predicted from the single fitted constant
$f$, not fitted per row.}
\label{tab:app_speedup_ladder}
\setlength{\tabcolsep}{3.6pt}

\begin{tabularx}{\linewidth}{@{}>{\raggedright\arraybackslash}p{1.5cm}
*{6}{>{\centering\arraybackslash}X} @{}}
\toprule
Backbone & Skip & $\rho$ & $S_{\mathrm{theo}}$ & $S_{\mathrm{meas}}$ &
$S_{\mathrm{pred}}$ & $n$ \\
\midrule
Wan2.1 & 50 & 0.50 & 1.935 & 1.877 & 1.879 & 8 \\
 & 52 & 0.52 & 2.011 & 1.947 & 1.947 & 31 \\
 & 54 & 0.54 & 2.092 & 2.021 & 2.021 & 19 \\
 & 56 & 0.56 & 2.180 & 2.098 & 2.100 & 21 \\
 & 58 & 0.58 & 2.276 & 2.184 & 2.186 & 28 \\
 & 60 & 0.60 & 2.381 & 2.279 & 2.279 & 39 \\
 & 64 & 0.64 & 2.622 & 2.492 & 2.492 & 39 \\
 & 66 & 0.66 & 2.762 & 2.613 & 2.614 & 18 \\
 & 68 & 0.68 & 2.918 & 2.749 & 2.748 & 36 \\
 & 70 & 0.70 & 3.093 & 2.897 & 2.897 & 16 \\
 & 72 & 0.72 & 3.289 & 3.065 & 3.063 & 47 \\
 & 74 & 0.74 & 3.513 & 3.251 & 3.249 & 18 \\
 & 76 & 0.76 & 3.769 & 3.463 & 3.459 & 48 \\
 & 78 & 0.78 & 4.065 & 3.703 & 3.699 & 64 \\
 & 80 & 0.80 & 4.412 & 3.980 & 3.974 & 32 \\
\midrule
FLUX & 28 & 0.56 & 2.223 & 2.195 & 2.197 & 10 \\
 & 31 & 0.62 & 2.558 & 2.521 & 2.521 & 17 \\
 & 32 & 0.64 & 2.694 & 2.650 & 2.651 & 25 \\
 & 34 & 0.68 & 3.013 & 2.957 & 2.956 & 42 \\
 & 35 & 0.70 & 3.202 & 3.138 & 3.136 & 18 \\
 & 36 & 0.72 & 3.417 & 3.340 & 3.340 & 36 \\
 & 37 & 0.74 & 3.663 & 3.573 & 3.573 & 56 \\
 & 38 & 0.76 & 3.947 & 3.842 & 3.840 & 33 \\
 & 39 & 0.78 & 4.279 & 4.147 & 4.150 & 36 \\
 & 40 & 0.80 & 4.672 & 4.514 & 4.514 & 55 \\
 & 41 & 0.82 & 5.144 & 4.947 & 4.949 & 73 \\
 & 42 & 0.84 & 5.723 & 5.475 & 5.476 & 49 \\
 & 43 & 0.86 & 6.448 & 6.124 & 6.130 & 16 \\
\bottomrule
\end{tabularx}
\vspace{-2mm}
\end{table}

\subsection{Analytic Cost Model and Its Validation}
\label{app:cost_model}

Table~\ref{tab:app_cost_model} gives the per-forward cost derived from
the architecture specification.
A skipped step is not free --- it still runs the probe (one
transformer block), which costs $3.3\%$ of a full forward on Wan2.1
and $1.8\%$ on FLUX.
FLOPs count attention and FFN matrix multiplies only and exclude the
text encoder and VAE, which are identical across all methods and do
not vary with the skip count.

Because the excluded components are fixed costs, measured speedup must
fall short of the analytic value.
Fitting the single-parameter relation
\begin{equation}
\frac{1}{S_{\mathrm{meas}}}
=
f+\frac{1-f}{S_{\mathrm{theo}}}
\label{eq:app_overhead_fit}
\end{equation}
independently at every operating point recovers an $f$ that varies
little across operating points:
$f{=}0.0323$ on Wan2.1 (median over $15$ operating points spanning
skip $50$--$80$; per-point range $0.0318$--$0.0333$; implied speedup
ceiling $1/f{=}31\times$) and
$f{=}0.0095$ on FLUX (median over $13$ points spanning skip
$28$--$43$; range $0.0092$--$0.0103$; ceiling $105\times$).
Substituting the median $f$ back reproduces every measured speedup to
within $0.16\%$ on Wan2.1 and $0.09\%$ on FLUX (worst case over the
$28$ operating points), so the shortfall is a fixed overhead, not a
breakdown of the cost model at high skip rates.
Table~\ref{tab:app_speedup_ladder} lists the full ladder; skip counts
with fewer than eight prompts are excluded from the fit
(Wan2.1: $46,48,49,53,55,57,62,63,67$; FLUX: $27,29,30,33$).

\subsection{Codec-Independence Check}
\label{app:codec}

Every video quality number in this paper is computed on H.264-encoded
output, so we recomputed the entire paired analysis from the raw
decoder tensors (lossless) on the $32$ held-out Wan2.1 prompts.
The re-generation needed for this check reproduced the original
outputs byte-for-byte in $192/192$ cases, so the lossless-versus-lossy
gap below is codec error alone and not run-to-run nondeterminism.

\begin{table}[t]
\centering
\small
\caption{\textbf{Codec-independence check on Wan2.1 ($n{=}32$ held-out
prompts): median paired differences recomputed from raw decoder
tensors (lossless) versus the H.264 pipeline (lossy).} All $18$ cells
keep the same adjudication.}
\label{tab:app_codec}
\setlength{\tabcolsep}{2.6pt}
\resizebox{\columnwidth}{!}{
\begin{tabular}{lllccc}
\toprule
Level & Pair & Metric ($n$) & Lossless $\tilde{\Delta}$ &
Lossy $\tilde{\Delta}$ & Same verdict? \\
\midrule
Moderate & M vs.\ D & LPIPS (29) & $-0.02773$ & $-0.02733$ & \checkmark \\
Moderate & M vs.\ D & PSNR (29) & $-3.55151$ & $-3.48327$ & \checkmark \\
Moderate & M vs.\ D & SSIM (29) & $-0.02829$ & $-0.02752$ & \checkmark \\
Moderate & S vs.\ D & LPIPS (31) & $-0.03436$ & $-0.03353$ & \checkmark \\
Moderate & S vs.\ D & PSNR (31) & $-2.84926$ & $-2.72569$ & \checkmark \\
Moderate & S vs.\ D & SSIM (31) & $-0.01354$ & $-0.01244$ & \checkmark \\
Moderate & S vs.\ M & LPIPS (30) & $-0.00504$ & $-0.00515$ & \checkmark \\
Moderate & S vs.\ M & PSNR (30) & $-0.04206$ & $-0.01858$ & \checkmark \\
Moderate & S vs.\ M & SSIM (30) & $+0.01219$ & $+0.01222$ & \checkmark \\
Fast & M vs.\ D & LPIPS (29) & $+0.02048$ & $+0.00976$ & \checkmark \\
Fast & M vs.\ D & PSNR (29) & $+1.15731$ & $+1.12510$ & \checkmark \\
Fast & M vs.\ D & SSIM (29) & $+0.13101$ & $+0.12565$ & \checkmark \\
Fast & S vs.\ D & LPIPS (29) & $-0.05325$ & $-0.05408$ & \checkmark \\
Fast & S vs.\ D & PSNR (29) & $-0.63552$ & $-0.63893$ & \checkmark \\
Fast & S vs.\ D & SSIM (29) & $+0.02509$ & $+0.02333$ & \checkmark \\
Fast & S vs.\ M & LPIPS (32) & $-0.06922$ & $-0.06335$ & \checkmark \\
Fast & S vs.\ M & PSNR (32) & $-1.74386$ & $-1.69897$ & \checkmark \\
Fast & S vs.\ M & SSIM (32) & $-0.11371$ & $-0.11071$ & \checkmark \\
\bottomrule
\end{tabular}
}
\vspace{-2mm}
\end{table}

\begin{table}[t]
\centering
\small
\caption{\textbf{Operating points of the two released default FLUX baseline
configurations on the same held-out set ($n{=}32$), with our
matched-compute levels for reference.} ``Worst'' is the maximum
per-prompt LPIPS.}
\label{tab:app_released_defaults}
\setlength{\tabcolsep}{2.4pt}
\resizebox{\columnwidth}{!}{
\begin{tabular}{llcccccc}
\toprule
Configuration & Signal & $\tau$ & $\lfloor \mathrm{ret}\cdot N\rfloor$ &
$\bar{s}$ & LPIPS$\downarrow$ & Worst & Speedup \\
\midrule
Official default (M) & $|\Delta y{-}\Delta x|$ & 0.08 & 0 & 41 & 0.2496 & 0.4736 & 4.94$\times$ \\
Official default (D) & $\Delta y$ & 0.4 & 10 & 35 & 0.1680 & 0.3401 & 3.13$\times$ \\
\midrule
Ours Moderate & S & 4.6 & 5 & 34 & 0.0816 & 0.2048 & 2.95$\times$ \\
Ours Moderate & D & 0.15 & 5 & 34 & 0.1353 & 0.3182 & 2.95$\times$ \\
Ours Moderate & M & 0.042 & 5 & 34 & 0.1014 & 0.2681 & 2.95$\times$ \\
Ours Fast & S & 6.4 & 5 & 37 & 0.1246 & 0.2487 & 3.57$\times$ \\
Ours Fast & D & 0.25 & 5 & 37 & 0.1842 & 0.3649 & 3.60$\times$ \\
Ours Fast & M & 0.065 & 5 & 37 & 0.1474 & 0.3866 & 3.57$\times$ \\
\bottomrule
\end{tabular}
}
\vspace{-4mm}
\end{table}

Compression error does not fully cancel in a paired difference.
In the worst LPIPS cell (Fast, M vs.\ D) the codec moves the paired
median by $0.0107$, i.e.\ $52\%$ of that cell's effect size
($0.0205$); the verdicts agree, but the margin is not uniformly large,
and we report that cell rather than the average.
Re-deriving every verdict from raw tensors leaves all $18$ judgments
unchanged.

\subsection{Released-Default Operating Points}
\label{app:released_defaults}

A calibrated-threshold comparison invites the objection that the
baseline was run at an unfavourable operating point.
Table~\ref{tab:app_released_defaults} therefore places the two
released default configurations of the FLUX baseline on the same
held-out set.
These rows carry no significance test: their skip counts differ from
each other and from our matched-compute levels, so a direct comparison
would confound quality with compute.
They are reported to show that our calibration does not weaken the
baseline: the official M default lands at a \emph{more aggressive}
operating point (median skip $41$) than either of our levels
($34$ and $37$).

The fourth column is the released threshold
$\lfloor \mathrm{ret}\cdot N\rfloor$, a step index rather than a count
of non-skippable forwards: FLUX holds a step exact when the counter is
at most this index (inclusive, hence one more forward than the index)
and additionally forces the final step, so the actual non-skippable
count is two larger.

\ifSuppStandalone
\bibliography{main}
\end{document}
\fi

\end{document}